\documentclass{article}

\usepackage[final,nonatbib]{neurips_2021}
\usepackage[numbers]{natbib}

\usepackage[utf8]{inputenc} % allow utf-8 input
\usepackage{microtype}
\usepackage{graphicx}
\usepackage{amsthm}
\usepackage{amssymb}
\usepackage{amsmath}
\usepackage{amstext}
\usepackage{wrapfig}
\usepackage{nicefrac}   
\usepackage{bm}
\usepackage{mathtools}
\usepackage{upgreek}
\usepackage{graphicx}
\usepackage{subcaption}
\usepackage{multirow}
\usepackage{makecell}
\usepackage{multicol}
\usepackage{algorithm}
\usepackage{algorithmic}
\usepackage{stackengine}
\usepackage{enumitem}
\usepackage{booktabs}
\usepackage{tablefootnote}
\usepackage{color, colortbl}
\usepackage{wrapfig}
\usepackage{setspace}
\usepackage{tkz-euclide}
\usepackage{etoolbox}
\usepackage{bbm}
\usepackage{breqn}
\usepackage{hyperref}
\usepackage{xcolor}

\newtheorem{theorem}{Theorem}[section]

\newtheorem{des}[theorem]{Desiderata}

\DeclareMathOperator*{\argmin}{argmin}
\DeclareMathOperator*{\argmax}{argmax}

\def \vx {\mathbf{x}}   % vector x, single image
\def \vy {\mathbf{y}}   % vector y, single image
\author{%
  Tianshi Cao $^{1,2,3}$ \thanks{Equal contribution} \hspace{0.8cm}
  Sasha Doubov $^{1,2}$ \footnotemark[1] \hspace{0.8cm}
  David Acuna $^{1,2,3}$\hspace{0.8cm} 
  Sanja Fidler $^{1,2,3}$\\
  \small{University of Toronto$^1$\quad Vector Institute$^2$\quad NVIDIA$^3$}\\
  \texttt{\{jcao,doubovs,davidj,fidler\}@cs.toronto.edu} \\
}

\title{Scalable Neural Data Server: A Data Recommender for Transfer Learning}
\begin{document}

\maketitle

\begin{abstract}
 Absence of large-scale labeled data in the practitioner's target domain can be a bottleneck to applying machine learning algorithms in practice.
 Transfer learning is a popular strategy for leveraging additional data to improve the downstream performance, but finding the most relevant data to transfer from can be challenging. Neural Data Server (NDS)~\cite{yan2020neural}, a search engine that recommends relevant data for a given downstream task, has been previously proposed to address this problem. % (Yan et al., 2020).
 NDS uses a mixture of experts trained on data sources to estimate similarity between each source and the downstream task.
 Thus, the computational cost to each user grows with the number of sources. % and requires an expensive training step for each data provider.
 To address these issues, we propose Scalable Neural Data Server (SNDS), a large-scale search engine that can theoretically index thousands of datasets to serve relevant ML data to end users.
 SNDS trains the mixture of experts on intermediary datasets during initialization, and represents both data sources and downstream tasks by their proximity to the intermediary datasets.
 As such, computational cost incurred by SNDS users  remains fixed as new datasets are added to the server. %, without pre-training for the data providers.
 We validate SNDS on a plethora of real world tasks and find that data recommended by SNDS improves downstream task performance over baselines. We also demonstrate the scalability of SNDS by showing its ability to select relevant data for transfer outside of the natural image setting.

\end{abstract}

\vspace{-3mm}
\section{Introduction}\label{sec:intro}
In recent years, machine learning (ML) applications have taken a foothold in many fields. Methodological and computational advances have shown a trend in performance improvements  achieved by using larger and larger training datasets, confirming that data is the fuel of modern machine learning. This fuel, however, is not free of cost. While the raw data itself (such as images, text, etc.) can be relatively easy to  collect, annotating the data is a labour intensive endeavour. A popular and effective approach to reduce the need of labeling in a target application domain is to leverage existing datasets via transfer learning. %This has been shown to be an effective strategy to boost performance in a target domain
Transfer learning is the re-purposing of ML models trained on a source dataset towards a different target task \cite{goodfellow2016deep}. 
The performance of transfer learning is predicated on the relevance of the source to the target \cite{taskonomy2018, raghu2019transfusion}. 
Although there are numerous datasets available through various data sharing platforms \citep{visualio, aimultiple2020,Snowflake2020,fysical2020}, finding the right dataset that will most benefit transfer-learning performance on the target domain is not a simple problem.

%In our work, we envision a large \emph{ML data marketplace} where data providers....
%\emph{Data Marketplaces} are platforms that allow data to be shared, traded and monetized such that their value is fully realized. A plethora of data marketplaces have been recently proposed or established, which allow individuals or businesses to monetize their own data, or purchase other's data \citep{aimultiple2020,Snowflake2020,Datum2020,fysical2020}. 

%A major challenge limiting the use of data marketplaces is the difficulty of finding useful data.
% \tianshi{although this example is not essential to our story, it kind of explains the question of "why not just search for tags/keywords".}

%\begin{wrapfigure}{R}{0.4\textwidth}
\begin{figure*}[t!]
    % \centering
    \vspace{-4mm}
    \begin{minipage}{0.59\linewidth}
    \includegraphics[width=0.98\textwidth]{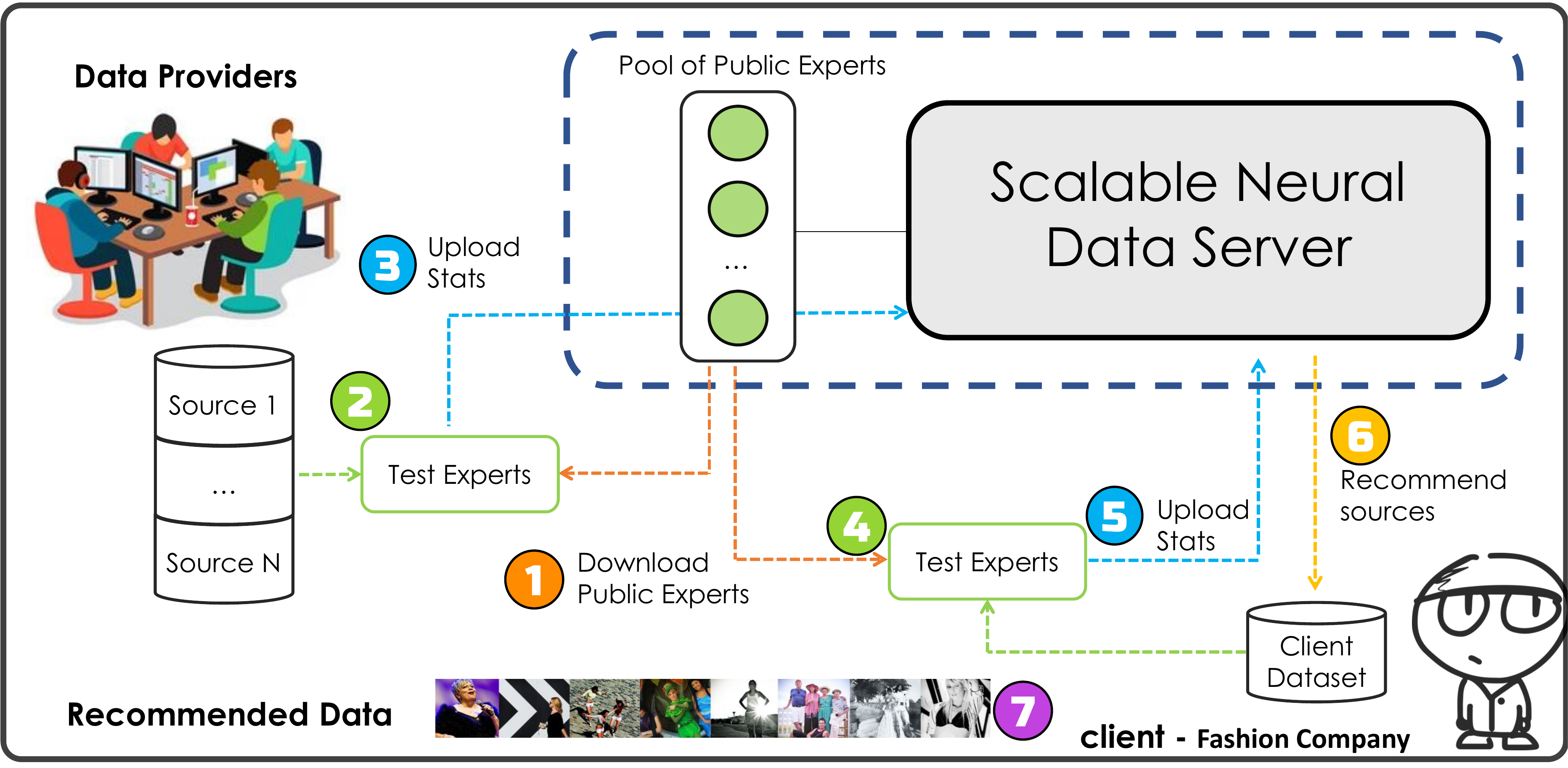}
    \end{minipage} \begin{minipage}{0.405\linewidth}
    \caption{\small\textbf{Scalable Neural Data Server.} SNDS is a scalable data recommendation system for transfer learning data. It can 
     theoretically 
    index thousands of datasets by using a \emph{pool of experts}, trained once on intermediary datasets, and used for data matching. SNDS uses these intermediary datasets to compute similarities between the consumer's data and the server's  datasets. Similarities are computed on the user's side with a computation cost that does not grow with the size of the server's data. User's data privacy is preserved. 
    % Key to this approach is a \emph{pool of experts}, trained only once on public datasets, and used for data matching. \sasha{TODO: remove one of figure 1 or 3}
    % by  data providers and data consumers.
    % to serve relevant ML data to endusers Experts trained on public datasets are used by both data providers and data consumers to process their dataset for matching. %SNDS indexes source datasets and the client dataset using a pool of experts, and recommends matching sources to the client.}
    }
    \label{fig:teaser}
    \end{minipage}
    \vspace{-6mm}
%\end{wrapfigure}
\end{figure*}

% \sasha{TODO: shorten example}
% For example, consider a startup looking to train a debris detector for use on a road cleaning robot, and suppose that they have collected a small, labeled dataset. They are looking for an abundance of data with which to augment their training set. What data should the startup use? Ideally, they would find a dataset of roadside debris that is captured from a sensor similar to their own. If they search for ``roadside debris", they would likely find data geared towards autonomous driving, which might be from one of the many sensor types (rgb, lidar, radar) and representations (bird's eye view, driver's view). Or, if they search for their specific sensor configuration, they might obtain data captured by similar setups, but of very different objects. 
% While keyword based searching can help filter out unrelated data, one is still faced with the problem of comparing apples to oranges. %What is needed is a recommendation system for finding the most relevant data to a target task.
% In this work, we envision a data search engine that indexes source data from data providers, and recommends relevant transfer learning data to consumers who have a target task. 

Consider a startup looking to train a debris detector for use on a road cleaning robot, and suppose that they have collected a small, labeled dataset. They are looking for data with which to augment their training set. %What data should the startup use? 
Ideally, they would find a dataset of roadside debris that is captured from a sensor similar to their own. However, keyword searches for ``roadside debris", would likely find data geared towards autonomous driving, which might be from one of the many sensor types (rgb, lidar, radar). % and representations (bird's eye view, driver's view). 
Searching for their specific sensor type might lead to data from similar setups, but of different objects. Rather than using keywords, an ideal system would find relevant data based on the \textit{training set} itself. In this work, we envision a data search engine that indexes source data from data providers, and recommends relevant transfer learning data to consumers who have a target task.
%Or, if they search for their specific sensor configuration, they might obtain data captured by similar setups, but of very different objects. 
% While keyword based searching can help filter out unrelated data, one is still faced with the problem of comparing apples to oranges. %What is needed is a recommendation system for finding the most relevant data to a target task.

Neural DataServer (NDS) \cite{yan2020neural} has been proposed as a data recommendation system for transfer learning. Given a target task (as represented by a target dataset), it uses expert models trained on splits of source data to make predictions on the target, which is in turn used as a measure of similarity between the source and the target. %Unlike the setup considered in NDS, in which the number of partitions is small and controlled by the server, the number of sources in a data search engine grows over time and can be very large.
Directly indexing a large number of datasets with NDS is expensive, as it would require each data provider to upload a trained expert, and any data consumer looking to select data for purchase must download and run all of the trained experts in order to evaluate similarity. Hence, for $M$ data providers in the server, the computational and bandwidth cost to the data consumer scales linearly with $M$.

% \sasha{For $M$ data providers and $N$ data consumers, %the computational capacity of the full system is $O(M+N)$, but 
% the computational cost scales by $O(MN)$.}

% \sasha{There is also privacy leakage in NDS. When experts are trained, they may memorize certain examples in the source. It has been shown that is possible to reconstruct training images from trained CNNs by optimizing images from noisy priors \cite{fredrikson2015model, zhang2020secret, chen2020improved}. Thus, an ill-behaving data consumer could use the source-trained experts to obtain data from the sources without purchasing/licensing it from the data-provider. A privacy preserving solution should treat Neural Networks (NN) with the same caution afforded to its training data.}

At the same time, using an expert model per data provider in NDS ensures that the data recommendations are highly relevant to the consumer. A data search engine that operates in a large-scale setting must be able to preserve the quality of recommendations, while disentangling the consumer's computational cost from the number of providers. The system should also preserve the data privacy properties from NDS, where the client's data is never exchanged with the server. % during the recommendation process. %As well, the system should ideally be able to retrieve relevant data  

We propose Scalable Neural Data Server (SNDS), a search engine for data recommendation at scale.  
This is achieved by using an intermediary, public dataset to train a fixed number of experts, and a similarity measure that evaluates the relevance of the server's data for a given consumer via these experts. Compared to NDS, the computational cost of SNDS is constant for a given consumer with respect to the $M$ data providers, and so is the bandwidth cost between the consumer and the server. We validate SNDS experimentally and find that 1) SNDS generalizes in the sense that it recommends useful data for downstream tasks that are not directly represented by the intermediary datasets, 2) SNDS improves performance on downstream tasks over baselines, and is competitive with NDS, 3) SNDS is able to generalize to domains that are very dissimilar to the intermediary datasets.

\vspace{-0mm}
\section{Background} \label{sec:background}
\vspace{-0mm}

In this section, we first formalize our problem setting which is similar to that of NDS, and then provide a brief overview of the data selection method introduced in NDS.

\vspace{-2mm}
\subsection{Problem setting} 
Let $\mathcal{X}$ denote a sample space and $\mathcal{Y}$ a label space. Suppose that there are currently $M$ data providers (which we call ``sources"), each with data available for use in transfer learning. Let the sources be denoted as $\mathbb{S} = \{S_1, \dots, S_M\}$, where $S_i$ denote a set of $m_i$ sample-label pairs $((\vx_1, \vy_1), \dots, (\vx_{m_i}, \vy_{m_i})) \in (\mathcal{X} \times \mathcal{Y})^{m_i}$. 
The target task by the data consumer is represented by a target specific distribution $\mathcal{D}$ supported over $\mathcal{X} \times \mathcal{Y}$, from which the target dataset $T = ((\vx_1, \vy_1), \dots, (\vx_{n}, \vy_{n})) \in (\mathcal{X} \times \mathcal{Y})^{n}$ is drawn. 
%\DA{I am not sure about that notation. I have seen $(x,y)^n \subset \mathcal{X} \times \mathcal{Y}$. but this is not the same, it gives the impression of repeating the same datapoint n times?.}
The goal of the data consumer is to find data sources suitable for transfer learning to their target task. 
Specifically, we seek a mixture of sources $S^*$ that minimizes the following risk for the data consumer:
\begin{equation}\label{eq:goal}
    R(S) = E_{\vx,\vy \sim \mathcal{D}}[l(h_{S \cup T}(\vx), \vy)]
\end{equation}
while satisfying some budget constraint of $|S| \leq b$, as specified by the data consumer. Here, $h_{S \cup T}$ is a predictive model $h: \mathcal{X} \rightarrow \mathcal{Y}$ that is trained on $S$ and $T$. In practice, $h$ is often first trained on $S$ during pretraining, and then trained on $T$ during finetuning. %\SF{S and T notation changes}
%\DA{the order matter, we should say sth like with S first in practice?}

%\DA{Should we formalize this more, For example, \textbf{Desiderata}. The search engine should satisfy... then D1, D2,D3,.... Easy to say later on, NDS does not satisfy D3 and D4}
% \paragraph{Desiderata} 
\begin{des}
\iffalse
To satisfy the requirements of a data search engine, we consider the following constraints to any potential solution:
\fi
To meet the requirements of a data search engine, a potential solution should satisfy:

    $\bullet$\ \textbf{D1}: Data in $\mathbb{S}$ or $T$ are not stored on the server.
    
    $\bullet$\ \textbf{D2}: Data in $\mathbb{S}$ is not revealed to the consumers before $S^*$ is selected.
    
   $\bullet$\ \textbf{D3}: Data in $T$ is not revealed to any data provider or any other consumers throughout the process.
   
    %$\bullet$\ \textbf{D4}: The per-query computational cost incurred by a consumer is no more than $O(n)$, that is, it is not more expensive asymptotically than iterating through their own dataset once. Similarly, the cost for each provider $i$ to index their data with the search engine is no more than $O(m_i)$. 
    
    $\bullet$\ \textbf{D4}: The computational and bandwidth cost for the consumer should be fixed as $M$ (i.e the number  of data providers) grows. 
    % \sasha{The computational cost of the consumer and server should be fixed with respect to the $M$ data providers. }
    %The 
   %\sasha{ $\bullet$\ \textbf{D5}: The total computational cost for indexing $M$ data sources and serving $N$ queries should be no more than $O(M+N)$.}

\end{des}
What makes this problem challenging is that D1-D3 prevents the sources and the target from being on the same device, and the D4 prevents the server from iterating through every example in $\mathbb{S}$ for responding to every data consumer. %Furthermore, D4 prohibits the computational and bandwidth cost between the server and the client to grow linearly with the number of data providers. 
%In the case where $N$ data consumers need to be served, we consider the task of recommending data to each consumer independently. In such case, we let $T = \{T_1,\dots, T_N\}$ denote the collection of $N$ targets, and let $n_j$ denote the size of target set $T_j$.
Our problem setting differs from that of NDS in that we consider the number of sources $M$ as an extrinsic variable, i.e. something outside the control of the server. This necessitates a solution that scales cheaply with the number of sources, as stated in D4.
\vspace{-2mm}
\subsection{Neural Data Server} 
\vspace{-1mm}
%\DA{Shouldnt this section go first? The formulation from 1 was introduced in NDS as well. }
%In this section, we provide a brief overview of Neural Data Server and explain why it is not suitable for application in the data marketplace.
NDS partitions the source data through either (1) superclass partition or (2) unsupervised partition. In either case, image features are extracted with a pretrained network. The image features are either (1) averaged over samples of the same class and then clustered with k-means, or (2) clustered with k-means directly. We refer to each of these clusters as $S_i$ ($i$ ranges from $1$ to $K$) as these are analogous to the sources considered in our problem formulation. Then, expert networks are trained for rotation prediction on each partition. Specifically, for each image $\vx$ and a rotation angle $\theta \in \{0, 90, 180, 270\}$, let $r(\vx,\theta)$ denote the image rotated clockwise by $\theta$. An expert is a function $f:\mathcal{X} \rightarrow \Delta^3$, where $\Delta^n$ is a probability mass function of $n+1$ choices.
The loss minimized by a expert $f_i$ is:
    $\mathcal{L} = -\sum_{\vx \in S_i} \sum_{j=0}^3 \log f_i(r(\vx,j))_j.$
    
Once an expert is trained, accuracy of image rotation prediction is evaluated on the target set:
\begin{equation}
    z_i = \frac{1}{4|S_i|}\sum_{\vx \in T} \sum_{j=0}^3 \mathbbm{1}(\arg \max_k[f_i(r(\vx,j))_k] == j).
\end{equation}
The importance weight $w_i$ for each partition $i$ is computed as $\textit{softmax}(\mathbf{z})_i$. The sample weight for each example in the source is computed as $\pi(\vx) = \sum_i^K \mathbbm{1}(\vx \in S_i)\frac{w_i}{|S_i|}$, which is used to sample examples without replacement from the union of source data sets.
Exposing the models directly trained on the source datasets in NDS to the client introduces privacy leakage, as the trained experts may memorize certain examples in the source data. It has been shown that is possible to reconstruct training images from trained CNNs by optimizing images from noisy priors \cite{fredrikson2015model, zhang2020secret, chen2020improved}.
We note that while NDS satisfies D1,
it does not satisfy D4 and  there may also be  privacy leak on D2.
% \DA{This is great! but I think it should be probably in the intro when we are contrasting (and probably in the intro of the method). In this section we only want to introduce what is need for the reader to understand SNDS without reading NDS}

\vspace{-2mm}
\section{Scalable Neural Data Server} \label{sec:snds}
\vspace{-1mm}
Scalable Neural Data Server (SNDS) is a data search engine for transfer learning. It is designed with application to large data sharing platforms in mind, where each data provider holds a source dataset. The server's goal is to find the most relevant data to a data consumer who has a budget constraint for the amount of data they could afford. 
%We also make explicit that the server performs recommendation over sources rather than over individual samples.
Since it is not feasible to enumerate through every example indexed by the server when responding to every consumer, we simplify the search problem from ``find the most relevant examples" to ``find the most relevant source datasets". This simplification was also implied by NDS, who uses experts to represent examples of each source split.
%Thus, $S$ can be generally expressed as $\bigcup_{i=1}^M Sample(S_i, w_i)$, where $Sample(S,w)$ is a function that returns $w$ fraction of $S$. The budget constraint is now defined as $\sum_{i=1}^M w_i|S_i| \leq b$, and our solution space to Eqn.~\ref{eq:goal} is simplified to the space of coefficients $\mathbf{w} \in [0,1]^M$.

SNDS is scalable in that the number of experts downloaded by each data consumer, as well as the amount of compute used in the transaction, remains constant regardless of the number of data providers. This satisfies the computational cost constraint, making it scalable to a large number of source datasets. SNDS is also privacy preserving for both the data consumer and the data providers: no source or target data, nor any models trained on either data, is moved until the recommendation is made. %This is in contrast to NDS, which is only privacy preserving for the data consumer. 
We introduce an intermediary public dataset for training a fixed set of experts. Outputs of these experts on source and target data are used as feature for matching sources to targets.

%\begin{wrapfigure}{l}{0.7\textwidth}
\begin{figure*}[t!]
\vspace{-2mm}
    \centering
    \begin{minipage}{0.72\linewidth}
    \includegraphics[width=0.98\textwidth]{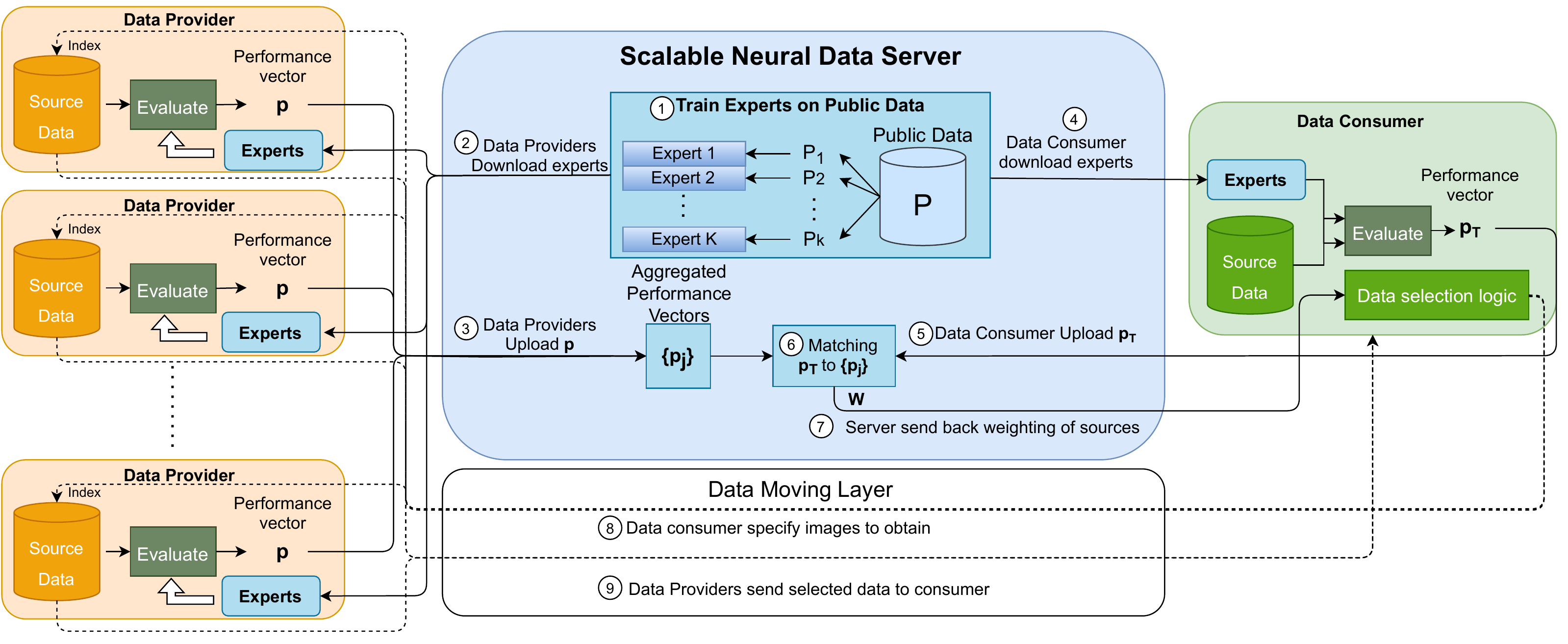}
    \end{minipage}
    \begin{minipage}{0.25\linewidth}
   \vspace{-1mm}
    \caption{\small Diagram of our Scalable Neural Data Server for recommending data to a single consumer.
    %Arrows indicate movement of information, and numbers indicate the order in which the process runs. 
    Step 1 occurs during initialization. Steps 2 \& 3 occur during indexing of data providers. Steps 4-7 occur during data selection. Steps 8-9 are data transactions external to SNDS.}
    \label{fig:snds_overview}
     \end{minipage}
    \vspace{-5mm}
\end{figure*}

\vspace{-2mm}
\subsection{SNDS Overview}
In SNDS, we assume that the server has access to a public dataset $P$. The server partitions $P$ into $K$ disjoint subsets, which we denote as $\{P_k\}_{k=1}^K$. The server then optimizes an expert $E$, belonging to some hypothesis class $\mathcal{E}$, for some objective $L$ with learning algorithm $\mathcal{A}_{L}$, for each $P_k$. We thereby obtain $K$ experts $\{E_k\}_{k=1}^K$.
The server defines a performance function $\mathcal{P}(D, E): \mathcal{X}^n \times \mathcal{E} \rightarrow \mathbb{R}$, which computes some metric using $E$ on a dataset $D$. The output of $\mathcal{P}$ is a single real number, but can also be vector valued. Each data provider $i$ runs the performance function on their data $S_i$. 
This provides a $K$ dimensional representation $\mathbf{p}_i = (p_{i,1}, \dots, p_{i,K})$ for each $S_i$ in $\mathbb{S}$, where $p_{i,k} = \mathcal{P}(S_i, E_k)$. 
At the same time, a data consumer also evaluates the experts on their target dataset $T$ with $\mathcal{P}$ to obtain a representation $\mathbf{p}_T = (p_{1,T}, \dots, p_{K,T})$. 
These representations are communicated back to the server, who then computes similarity between the target representation and that of each source. Let $sim(a,b):\mathbb{R}^K \times \mathbb{R}^K \rightarrow \mathbb{R}$ denote a similarity function. We then have $\mathbf{z} = (z_1, \dots, z_M |z_i = sim(\mathbf{p}_i, \mathbf{p}_T))$ as the similarity scores between the target and each source. Similarity scores are provided to the data consumer as the server's recommendation of sources. The data consumer may then use the similarity scores to select data sources for transfer learning.
% SNDS is illustrated in Fig~\ref{fig:snds_overview}.
%$\mathbf{z} = (z_1, \dots, z_M)$. Here, $z_i = \frac{||\mathbf{p}_i\cdot \mathbf{p}_T||^2}{||\mathbf{p}_i||||\mathbf{p}_T||}$ is the cosine similarity the representation of the target and that of source $i$. 

%Finally, we apply softmax with temperature on $\mathbf{z}$ to obtain weights $\mathbf{w} \in \mathbb{R}^M$ that sums up to one. Similar to NDS, we use $\mathbf{w}$ to weigh the sampling weight of each sample in $\mathbb{S}$, and sample from $\mathbb{S}$ without replacement. Indices of the selected data from each source is aggregated and suggested to the user. 

\vspace{-2mm}
\subsection{Server Initialization} \label{sec:setup}
\vspace{-1mm}

%The server needs to find data to use as the public dataset. As the name suggest, the content of the public dataset is not owned by any data provider, and ideally it should be in the public domain such that no privacy implication could arise from its use. Furthermore, this public dataset should be representative of the type of data found in the sources. %As such, one possible solution is for the server to require every data provider to contribute a small fraction of their data to this public dataset. 
%In our experiments, in the absence of data providers, we use the ImageNet Large Scale Visual Recognition Challenge (ILSVRC) 2012 training set as the public dataset. Image features are extracted by a pretrained ResNet34's penultimate layer \cite{he2016deep}. $K$-means is used to partition the ILSVRC2012 training split into 50 partitions. In deployment, an online approach could be also used, where the server starts with external datasets like ILSVRC2012 training split, and then add additional experts when more source data becomes available. 
In our experiments, we use the ImageNet Large Scale Visual Recognition Challenge (ILSVRC) 2012 training set as the public dataset. $K$-means on image features is used to partition the ILSVRC2012 training split into 50 partitions.
%The first step in setting up the server is to train experts on the public datasets.
Our main requirement for the experts is that their output as observed by $\mathcal{P}$ correlates with transfer performance from their partition of the public dataset. Following NDS, we use image rotation prediction as the task for training experts, as this task has been shown to correlate well with domain confusion, which is in turn an empirical estimate of transfer performance. As such, we train $E_k$ with loss function $L(P_k, E_k)$, where $L(D,E)=-\sum_{\vx \in D} \sum_{l=0}^3 \log E(r(\vx,l))_l$. % we restrict $\mathcal{E}$ to experts predicting rotation. 
Since experts only need to be trained once during initialization, the computational cost of this step does not contribute to the marginal cost of indexing or queries.

\vspace{-2mm}
\subsection{Indexing Sources} \label{sec:indexing}
\vspace{-1mm}

To add a source $S_i$ to SNDS, we evaluate experts $\{E_k\}_{k=1}^K$ on $S_i$ and store the results $\mathbf{p}_i$ on the server. First, the experts trained in section \ref{sec:setup} are downloaded by the data providers. Then, since the experts are trained for image rotation prediction, we use the rotation prediction accuracy as the evaluation metric. Specifically, $\mathcal{P}(D, E) = \frac{1}{4|D|}\sum_{\vx \in D} \sum_{j=0}^3 \mathbbm{1}(\arg \max_o[E(r(\vx,j))_o] == j)$. The representation of a source $S_i$ in the server is thus $\mathbf{p}_i=(\mathcal{P}(S_i, E_k))_{k=1}^K$. The computational cost of this step for data provider $i$ is $ \propto Km_i$, which satisfies the complexity constraint of $O(m)$. %Considering number of sources $M$ and average size of sources $m$ as asymptotic variables, the total asymptotic computational cost of this step is $O(Mm)$.

\vspace{-2mm}
\subsection{Data Selection}\label{sec:retrieval}
\vspace{-1mm}

A data consumer queries SNDS by first downloading experts $\{E_k\}_{k=1}^K$ and then evaluating each expert on their target dataset $T$.  Similar to Sec.~\ref{sec:indexing}, evaluation function $\mathcal{P}$ is used to obtain $\mathbf{p}_T = (\mathcal{P}(T, E_k))_{k=1}^K$, which is communicated back to the server. The asymptotic computational complexity of producing $\mathbf{p}_T$ is $O(n)$, which meets our requirements listed in Sec.~\ref{sec:background}. Furthermore, the number of downloaded experts and evaluations is independent of $M$ sources, and so SNDS satisfies D4 on the consumer-side.

The server needs to weigh each source. For simplicity and computational efficiency, we choose a normalized variant of cosine similarity because it is shift invariant in its input arguments and has a fixed output range. This alleviates the need to tune the selection strategy based on the performance metric. Specifically, we compute the channel mean as $\Bar{\mathbf{p}} = \frac{1}{M} \sum_{i=1}^M \mathbf{p}_i$, and use this to obtain centered performance vectors $\mathbf{p'}_{i} = \mathbf{p}_i - \Bar{\mathbf{p}}$ and $\mathbf{p'}_T = \mathbf{p}_T - \Bar{\mathbf{p}}$. We define the similarity function as $sim(\mathrm{a},\mathrm{b}) = \frac{||\mathrm{a}\cdot \mathrm{b}||^2}{||\mathrm{a}||||\mathrm{b}||}$, so that $\mathbf{z} = (\frac{||\mathbf{p'}_i\cdot \mathbf{p'}_T||^2}{||\mathbf{p'}_i||||\mathbf{p'}_T||})_{i=1}^M$. 

We normalize $\mathbf{z}$  via softmax with temperature $\tau$ to obtain source weights $\mathbf{w} \in \mathbb{R}^M$. The temperature parameter determines the trade-off between sampling greedily from the most relevant clusters and sampling uniformly from all clusters. Rather than assuming a constant value for $\tau$, we find the $\tau$ that satisfies a target entropy value for $\mathbf{w}$. Intuitively, this allows us to maintain the same trade-off between greedy vs. uniform sampling across different consumer datasets, and this approach outperforms a fixed  $\tau$ in practice. More details on this in the supplementary material.
\iffalse
We adjust $\tau$ with gradient descent, which stably converges within dozens of iterations. In all  experiments, we set the target entropy value to $1.5$ --  the first value we tried.
\fi
% This temperature parameter determines how sharp the weighing of the sources are. We adjust $\tau$ with gradient descent, using as objective the difference between the entropy of $\mathbf{w}$ and a fixed target value. This is a convex optimization problem that stably converges within dozens of iterations. In all  experiments, we set the target entropy value to $1.5$ --  the first value we tried.

Finally, the server outputs $\mathbf{w}$ to the data consumer, who uses it to select source data. 
%Computing $\mathrm{m}$ from $\mathbf{p}$s incurs an asymptotic cost of $O(MK)$ to the server.
Computing $\mathbf{w}$ from the expert scores incurs an asymptotic cost of $O(MK)$ to the server.
This is the main operation that scales with the number of data providers in the system. We note that millions of dot products between low (K) dimensional vectors can be performed in seconds on modern hardware, and the computational cost in both NDS/SNDS is vastly overshadowed by the  expert model evaluations on the consumer's data -- which is precisely the cost SNDS is trying to minimize.

\vspace{-2mm}
\subsection{Differences with NDS}
SNDS implements many similar techniques as NDS, but has significant differences to NDS conceptually. 
We highlight the main differences between SNDS and NDS:

    1. NDS is designed to find data from few datasets. SNDS is designed to find data from many datasets. 
    
    2. NDS trains an expert for each source. SNDS trains an expert for each intermediary public dataset.
    
    3. NDS uses performance of experts on the target dataset as proxy for similarity to sources. SNDS measures performance of (public) experts on sources and targets, and then uses similarity in performances as proxy for data similarity.
    
    4. Querying with a target dataset of size $n$ on NDS incurs a bandwidth cost $\propto M$ and computational cost $\propto M \times n\times C$. \
    Querying with a target dataset of size $n$ on SNDS incurs a bandwidth cost $\propto K$ and computational cost $\propto K \times n\times C$. Here $C$ is the cost of evaluating an expert on a data point. % $K$ is a constant while $N$ grows overtime. 

\vspace{-2mm}
\subsection{Theoretical Analysis}
The benefits of SNDS are achieved by disentangling the training of experts from sources: the computational and privacy drawbacks are due to the 1-to-1 correspondence between sources and experts. 
This ``transfer by proxy" approach works because intuitively, if tasks A and B are both ``similar" to C, then A and B are also ``similar" to each other. Indeed, theory in domain adaptation supports this intuition \cite{ben2007analysis, mansour2009domain, AcunaICML21_fDAL}. For example, \citet{mansour2009domain} have provided a generalization bound on the target domain.
% \DA{Shouldnt we put this as a subsection. i.e "Theoretical Analysis" and put it after 3.1? }

% \DA{For example. consider a version for method like this: . 
% \begin{itemize}
%     \item Method, 
%     \item 3.1. SNDS Overview 
%     \item 3.2 Server Side. 
%     \item 3.3 Client Side. 
%     \item 3.3 Difference vs NDS(This is not typical but i would do it because it is a follow up work and we want to show it is different enough) 
%     \item 3.4 Theoretical Analysis
% \end{itemize}}

\begin{theorem}
\citet{mansour2009domain}: Let $S$ and $T$ be the source and target domains over $\mathcal{X} \times \mathcal{Y}$, $\mathcal{H}$ be a hypothesis class, and let $l:\mathcal{Y} \times \mathcal{Y} \rightarrow \mathbb{R}^+$ be a symmetric loss function that obeys the triangle inequality. Further, let $h^*_S = \argmin_{h \in \mathcal{H}} R^l_S (h)$ and $h^*_T = \argmin_{h \in \mathcal{H}} R^l_T (h)$ denote the optimal hypothesis for their respective domains, we have
\begin{equation*}
    \forall h \in \mathcal{H}, R_T^l(h) \leq R_S^l(h, h^*_S) + disc_l(S_{\mathcal{X}}, T_{\mathcal{X}}) + R^l_T (h^*_T) + R^l_S(h^*_T,h^*_S),
\end{equation*}
where $R^l_S(h,h_S^*) = E_{\vx\sim S} \left[l(h(\vx), h^*_S(\vx))\right]$; $S_{\mathcal{X}}$ and $T_{\mathcal{X}}$ are source and target distributions marginalized over $\mathcal{X}$.
\end{theorem}

We make the simplifying assumption that the source task is the same as the target task, since it is not computationally feasible for the server to identify the task of every source and target. By the same vein, we assume that the risk on any $S$ is similar. Thus the main quantity we look to optimize is $disc_l(S_{\mathcal{X}}, T_{\mathcal{X}})$. Which is defined as:
\begin{equation}
    disc_l(S, T) = \sup_{(h,h') \in \mathcal{H}\times \mathcal{H}} \left|R^l_S(h',h) - R^l_T(h',h)  \right|
\end{equation}
An important property of $ disc_l$ is that it obeys triangle inequality independently of the choice of $l$. That is:  $disc_l(S,T) \leq disc_l(S,P) + disc_l(P,T)$ for some intermediate domain $P$ over the same subspace $\mathcal{X}$. A simple observation is then that if we minimize the upper bound, we are indeed minimizing the $disc(S,T)$. This observation fuels the motivation for our approach.

\vspace{-0mm}
\section{Experiments} \label{sec:experiments}
\vspace{-0mm}
%We first empirically verify that our choice of proxy task is well correlated with domain confusion, thereby allowing us use the triangle inequality between domain confusion to qualify similarity between source and target datasets. 

We first empirically verify that our similarity function output is well correlated with domain confusion, validating our choices of the proxy task and similarity function in the data selection process.
We then experimentally evaluate SNDS in the task of classification. We use accuracy of classifier trained through transfer learning on target datasets to gauge the usefulness of data recommended by SNDS. We also extend the experiments into domains dissimilar  to ImageNet, to test the generality of SNDS.% is able to be helpful in other areas.

\vspace{-0mm}
\subsection{Datasets} \label{sec:datasets}
\vspace{-0mm}
\paragraph{Source data} We simulate data providers with partitions of the OpenImages dataset\cite{OpenImages}. The OpenImages dataset contains 9 million images with various modes of annotations. We split OpenImages into 50 partitions using $K$-means on image features extracted by an ImageNet pretrained network. Each image in OpenImages is tagged with multiple object bounding boxes with labels. We convert OpenImages into a classification dataset by cropping the images by the bounding boxes and assigning the label of that bounding box as label of the cropped image. Thus, we obtain a classification dataset of 15.9 million cropped images, each with a label from the 600 object bounding box categories. %We resize all cropped images to a resolution of 256x256

\vspace{-0mm}
\paragraph{Public data and expert training}
We use the training split of ILSVRC2012\cite{ILSVRC15}, containing 1.2 million images, as the public dataset. We split it using superclass partition as described in \cite{yan2020neural}: the mean features of each class are extracted using an ImageNet pretrained network, and then clustered using $K$-means. Experts are trained on the task of image rotation prediction. Experts use ResNet18 as backbone with a input size of 224x224 and output dimension of 4 (corresponding to the 4 rotations). We use $K=50$ in our experiments, and study the effect of reducing $K$ in an ablation. Further details on transfer learning are found in the appendix. Experiments are implemented using PyTorch\cite{NEURIPS2019_9015}.

\vspace{-0mm}
\paragraph{Target Datasets}
We use nine finegrained classification datasets as target datasets. They are: FGVC-Aircraft \cite{maji13fine-grained}, Stanford Cars \cite{KrauseStarkDengFei-Fei_3DRR2013}, CUB200 \cite{WelinderEtal2010}, Stanford Dogs \cite{KhoslaYaoJayadevaprakashFeiFei_FGVC2011}, DTD \cite{cimpoi14describing}, Flowers102 \cite{Nilsback08}, Food100 \cite{bossard2014food}, Oxford Pets \cite{parkhi12a}, and SUN397 \cite{xiao2010sun}.

\vspace{-0mm}
\subsection{Domain confusion and SNDS}

\begin{wrapfigure}{R}{0.33\textwidth}
    \vspace{-18mm}
    % \centering
    \includegraphics[width=0.33\textwidth,trim=10 0 30 10,clip]{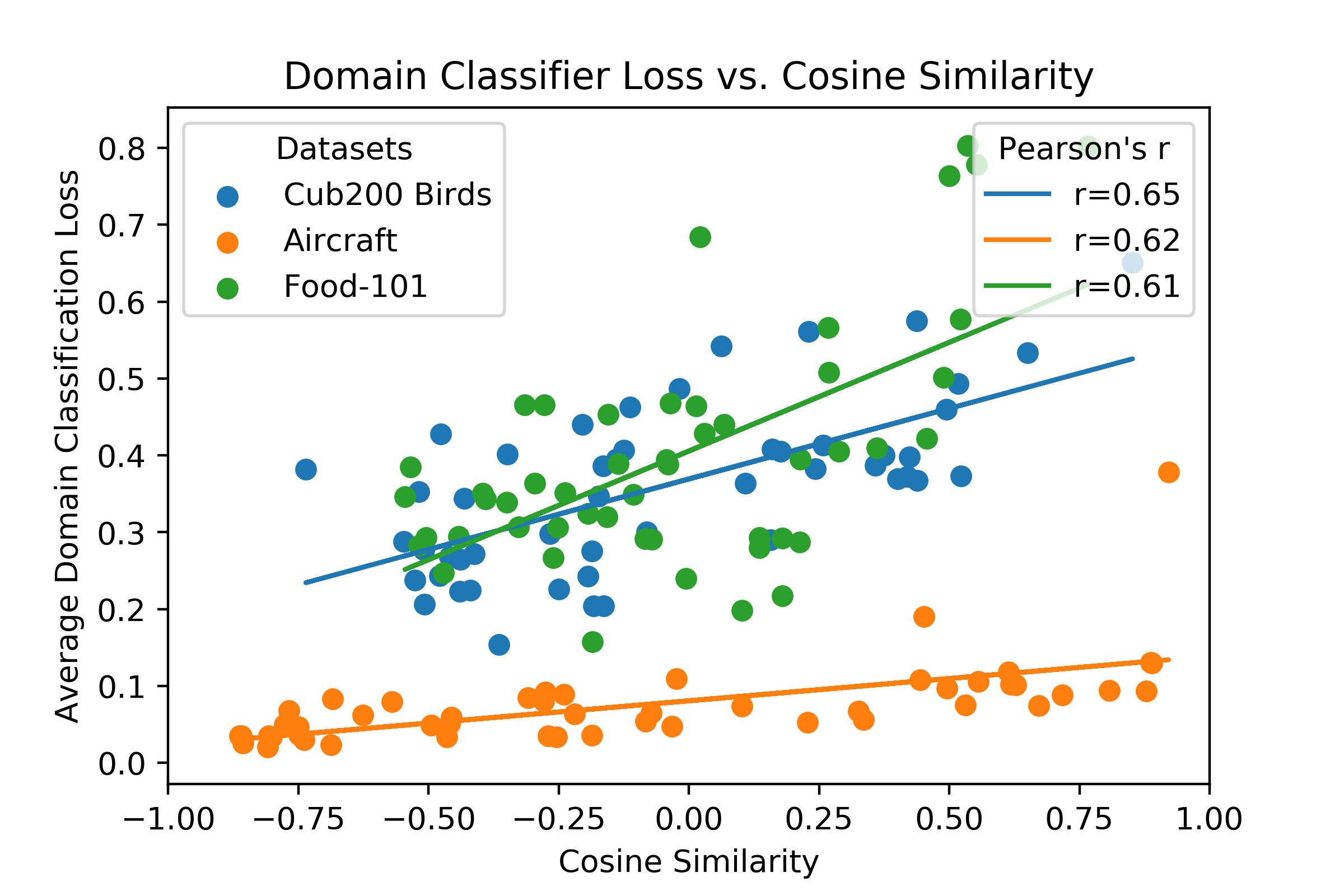}
    \vspace{-5mm}
    \caption{\small Relationship between the domain confusion and similarity metric of the source and target domains. }
    \label{fig:domain_confusion}
    \vspace{-12pt}
\end{wrapfigure}

%\tianshi{NDS already showed that $disc_l(A,C) \sim \mathcal{P}(E_A, C)$, where $E_A$ is expert trained on A, and C is target. We can experimentally demonstrate that $|disc_l(A,B)-disc_l(B,C)| \leq disc_l(A,C) \leq disc_l(A,B)+disc_l(B,C)$, and that $disc_l(A,C) \sim $ the proximity metric we use. It would be great if we can show that if we compute $|disc_l(A,B)-disc_l(B,C)|$ and $disc_l(A,B)+disc_l(B,C)$ enough times with different $B$s, then we can get a good approximation of $disc_l(A,C)$. But since triangle inequality itself isn't tight, we would need to find a better way to convey the: "N+1 unique $R^N$ spherical surfaces have at most one intersecting point" property.}
Since we want SNDS to select sources that are similar to the target domain, the scores of each source returned by SNDS should intuitively be correlated with $disc_l(S_i,T)$ for source $S_i$ and target $T$. Domain confusion \cite{ben2007analysis, ganin2016domain} has been used as an empirical approximation of $disc_l(A,C)$. In this experiment, we compare domain confusion and score provided by SNDS for 3 target datasets. 
%NDS have experimentally shown that $disc_l(A,C) \sim \mathcal{P}(E_A, C)$, where $E_A$ is an rotation prediction expert trained on partition $A$, and $C$ is the target domain. 
%However, the SNDS setting differs with the addition of the intermediary domain $B$, as the experts are not trained on source domain $A$ and rely on the similarity metric between $A$ and $C$ for data selection. Hence, we perform an experiment to compare the domain confusion and our computed similarity for domains $A$ and $C$.
%This is done by empirically estimating $disc_l(A_i,C)$ as done in NDS \sasha{should I cite other things aside from NDS} and computing $sim(A_i, C)$, where each $A_i$ is the source data of the $i$th data provider. \sasha{ordering of introducing OpenImages seems wrong}. 
%We use the partitions of OpenImages V6 for our source data and multiple fine-grained classification datasets for our target tasks and estimate $disc_l(A_i,C)$ with the average test loss of a classifer between domains $A_i$ and $C$. 
% Following the procedure of NDS, we use a ResNet18 network to classify between each pair of source and target. We sample 5k images from each domain for each epoch during training and a fixed sample of 2.5k images from each domain during testing.

We use an MLP with 1 hidden layer of 512 neurons to classify between each pair of source and target. We use a weighted sample of 50k images from the mixed source/target dataset to ensure that the samples are balanced, as done in NDS. We report the test loss on a held-out set of images from the source and target domains.  

As seen in Figure \ref{fig:domain_confusion}, the score returned by SNDS is positively correlated with the domain confusion between the source and target domains. This shows that SNDS indeed outputs higher scores for sources similar to the target domain.
This can also be observed qualitatively in Figure~\ref{fig:qualitative_figure}.
%despite using experts trained on an intermediary domain.  
%Source data partitions with a high cosine similarity (i.e. those most likely to be selected by SNDS) have high domain confusion which validates the use of our metric and the associated proxy task for data selection.

\vspace{-0mm}
\subsection{Data Recommendation with SNDS} \label{sec:dataRecommendation}

\vspace{-2mm}
\paragraph{Implementing Data Consumers} %We evaluate SNDS by performing transfer learning with data recommended by SNDS to downstream client tasks. 
Evaluation of SNDS on downstream tasks requires us to experimentally simulate data consumers. For target domain data, we use the finegrained classification tasks described earlier - each classification task represents one data consumer.
Once a recommendation of sources has been made by the server, it is up to the data consumer to decide on their budget, data selection strategy, and transfer learning method. 
Similar to NDS, our simulated data consumer sample from all available data in $\mathbb{S}$ without replacement, weighing the likelihood of each data point as $\pi(\vx) = \sum_i^M \mathbbm{1}(\vx \in S_i)\frac{\mathbf{w}_i}{|S_i|}$. 
%We discuss the impact of sampling strategy later in this section.
We pre-train on the selected data using supervised learning, and then finetune on the downstream dataset. Details about our transfer learning procedure are in the appendix.
Experiments are performed on a Tesla P100 with 12 GB memory in an internal cluster. %, available through Vector Institute's compute cloud.

\begin{table*}[t!]
    \centering
    \small
    \caption{Downstream task performance with pretraining data recommended by SNDS.}
    \vspace{-2mm}
    \resizebox{\linewidth}{!}{\begin{tabular}{lccccccccccc}
    \toprule
    \multirow{2}{*}{Selection Method} & \multirow{2}{*}{$\%$ Images} & \multicolumn{9}{c}{Target Dataset} & \multirow{2}{*}{\shortstack{Average}}\\
    & & CUB200 & Flowers102 & Pets & Food 101 & Stanford Dogs & DTD & Cars & Aircraft & SUN397 & \\
    \midrule
    No pretraining & $0\%$ &27.25 & 52.42 & 42.21 & 71.75 & 39.35 & 30.16 & 18.78 & 45.80 & 31.80 & 39.95 \\
    \midrule
    Random Sample & $2\%$ (292K)    & 41.49 & 74.03 & 67.02 & 72.34 & 52.93 & 52.72 & 46.77 & 54.66 & 31.80 & 55.75 \\
    SNDS-50 &          $2\%$ (292K) & 50.59 & 81.66 & 71.06 & 73.73 & 55.07 & 56.28 & 47.37 & 55.91 & 39.79 & 59.07 \\
    NDS & $2\%$ (292K)    & 49.91 & 78.32 & 70.71 & 73.35 & 54.64 & 55.32 & 52.95 & 57.95 & 42.27 & 59.53 \\
    \midrule
    Random Sample & $5\%$ (730K)   & 53.11 & 83.53 & 73.79 & 74.86 & 57.93 & 57.87 & 67.34 & 62.70 & 44.42 & 63.95 \\
    SNDS-50 &          $5\%$ (730K) & 57.35 & 87.68 & 75.91 & 74.87 & 59.38 & 60.59 & 63.80 & 62.32 & 43.99 & 64.96 \\
    NDS & $5\%$  (730K)   & 59.04 & 85.10 & 75.18 & 75.58 & 58.99 & 60.27 & 66.56 & 63.76 & 45.98 & 65.61 \\
    \midrule
    Random Sample & $10\%$  (1.46M)  & 58.40 & 86.82 & 76.40 & 76.22 & 60.81 & 60.74 & 71.58 & 65.24 & 47.69 & 67.10 \\
    SNDS-50 &       $10\%$ (1.46M)   & 61.02 & 89.89 & 77.32 & 77.29 & 61.33 & 64.41 & 72.68 & 65.12 & 45.89 & 68.33 \\
    NDS &  $10\%$ (1.46M)   & 61.27 & 89.88 & 78.32 & 77.03 & 61.28 & 60.96 & 73.77 & 65.57 & 46.84 & 68.33 \\
    % \midrule
    % Oracle (ImageNet) & - & 73.44 & 82.00 & 76.23	& 79.69 & 71.28 & 94.55 & 81.43 & 90.66 & 55.78 & 78.34 \\
    \bottomrule
    \end{tabular}}
    \label{tab:snds1}
    \vspace{-3mm}
\end{table*}

\begin{wrapfigure}{L}{0.58\textwidth}
\vspace{-9pt}
\captionof{table}{\small Downstream performance on Pets \& CUB200 Birds,  using pretraining data recommended by SNDS with no target class in public data}
\vspace{-2mm}
    \centering
    \small
    \resizebox{1\linewidth}{!}{
    \addtolength{\tabcolsep}{-3pt}
    \begin{tabular}[t]{l|c|c|c}
    \toprule
    \multirow{2}{*}{Selection Method} & \multirow{2}{*}{$\%$ Images} & \multicolumn{2}{c}{Target Dataset} \\
    & & CUB200 & Oxford Pets \\
    \midrule
    Random Sample & $2\%$ (292K)& 41.49 & 67.02 \\
    SNDS & $2\%$ (292K)& 50.59 & 71.06 \\
    SNDS (no target data) & $2\%$ (292K)& 47.67 & 70.35 \\
    \midrule
    Random Sample & $5\%$ (730K) & 53.11 & 73.79 \\
    SNDS & $5\%$ (730K) & 57.35 & 75.91 \\
    SNDS (no target data) & $5\%$ (730K)& 58.69 & 76.83 \\
    \bottomrule
    \end{tabular}}
    \label{tab:ablate}
\vspace{-10pt}
\end{wrapfigure}

% \subsection{Data Recommendation using SNDS}
% \subsubsection{Datasets}
% We use partitions of OpenImagesV6 as sources, and use several finegrained classification datasets (CUB200 Birds, Oxford-IIT Pets, and Flowers 102) as targets. Training split of ILSVRC2012 is used as the public dataset. We split OpenImages into 50 partitions using K-means on image features extracted by an imagenet pretrained network. We also split the public data into 50 partitions using the same method. 

% Each image in OpenImages is tagged with multiple image attributes, and object bounding boxes with labels are provided. Since all downstream tasks are that of classification, we convert OpenImages into a classification dataset by cropping the images by the bounding boxes and assigning the label of that bounding box as label of the cropped image. Thus, we obtain a classification dataset of 16 million cropped images, each labeled as one of the 600 categories.

\subsubsection{Main Experimental Results}\label{sec:mainExperiments}

In this experiment, we select $2\%$ (292K images), $5\%$ (730K images) and $10\%$ (1.46M images) of total source data using SNDS and compare with randomly selected data and NDS. We implement NDS with image-rotation experts, %and perform sampling without replacement
as per Sec.~\ref{sec:background}.
%We implement and test both $K=10$ and $K=50$ when splitting the public dataset with $K$-Means, and denote them by SNDS-10 and SNDS-50 in table~\ref{tab:snds1}. Using $K=50$ provides a more fine-grained splitting of the public dataset than $K=10$, thereby providing SNDS with experts that are more specialized to certain types of data.
Our results demonstrate that data selected by SNDS is more useful in almost every downstream task than randomly selected data. This advantage is most pronounced when using a tight budget of $2\%$, since the chance of randomly selecting useful data is higher when more data is being selected.

Furthermore, we find that SNDS is competitive with NDS even though NDS trains experts directly on source data, whereas SNDS trains experts on the public dataset. As such, SNDS promises cheaper computation cost with growing number of sources without sacrificing much in performance.

%\begin{wrapfigure}{R}{0.55\textwidth}
%\vspace{-10pt}
\begin{figure*}[t!]
    \centering
    \includegraphics[width=0.99\textwidth,trim=20 0 140 0,clip]{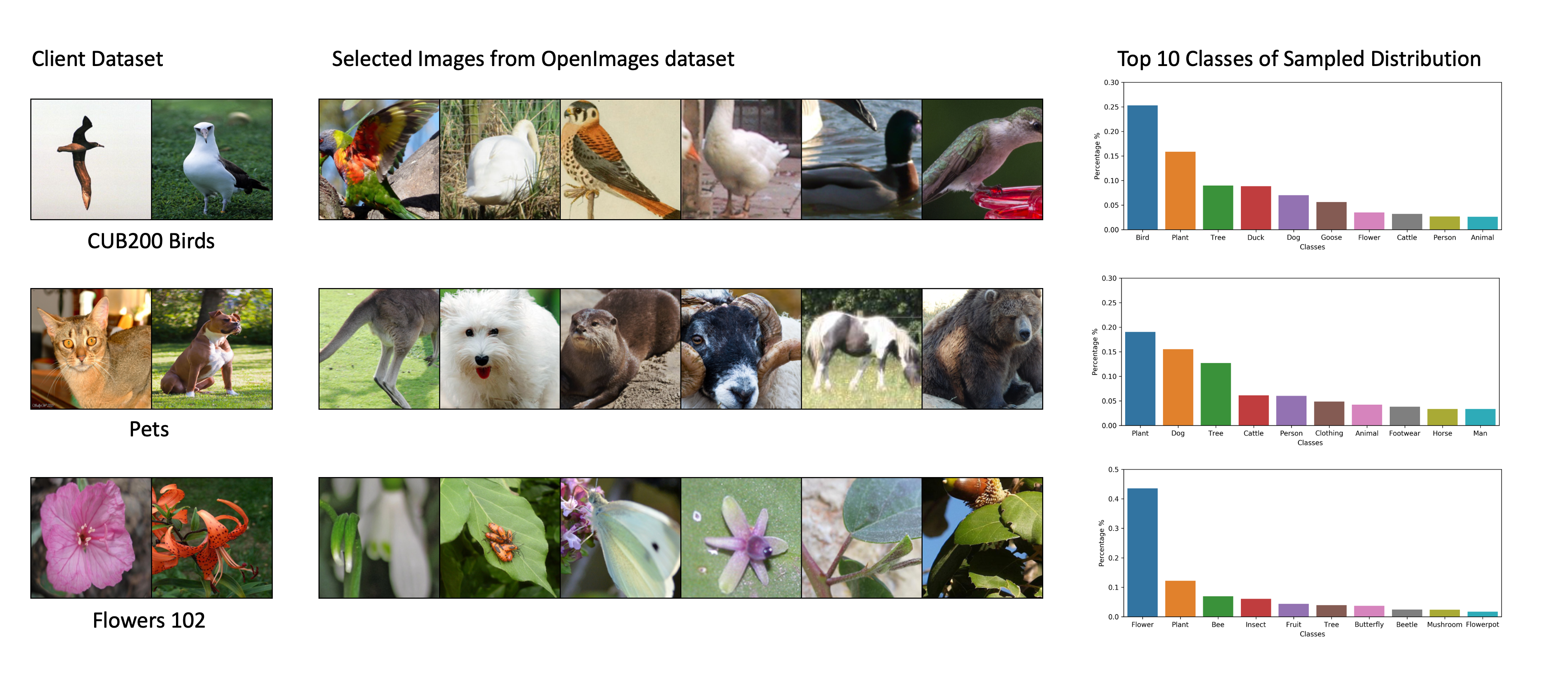}
    \vspace{-2mm}
    \caption{\small %Examples of selected source images from OpenImages and images from ``most-similar public split" from ImageNet for each client dataset, and the associated top classes for the sampled distribution.
    From left to right: Image from client dataset, images from ``most similar" public split (ImageNet), selected images from data sources (Openimages).
   }
    \label{fig:qualitative_figure}
%    \vspace{-15pt}
%\end{wrapfigure}
\vspace{-2mm}
\end{figure*}

Fig.~\ref{fig:qualitative_figure} shows  retrieval results from SNDS for several target datasets.  Similarity between the target and the public data split is determined by the accuracy of that split's expert on the target dataset. Since OpenImages provides multiple class labels per image, the top classes in the sampled distribution may be ones that co-occur with more relevant classes, such as the \textit{Plant} class for the Pets dataset. 

We ablate the use of 50 experts in the appendix, and find that with just 10 experts, SNDS is able to out-perform the baseline, though having more experts allows for more specific recommendations that improve the downstream performance.

%Furthermore, we find that SNDS-10, with just 10 experts, already achieves most of the advantage that NDS has over the baseline in the $2\%$ budget case. SNDS-50 further closes this gap and is
%competitive with NDS. This result shows that SNDS sacrifices very little performance but achieves better computational scaling in source data sets.

\vspace{-0mm}
\paragraph{Impact of missing public data} 
A critical factor determining the feasibility of SNDS is whether it generalize to types of images that are not represented in the public dataset. That is, it should be able to recommend source images similar to those in the target even if no such images are in the public dataset.
We emulate this by dropping images in the public dataset with classes corresponding to the target task. Two target tasks: Oxford Pets and CUB200, are used. After we split the public dataset into 50 superclass clusters, we remove clusters that contain overlap with target domain classes. For the Oxford Pets experiment, we removed eight clusters that included images containing dogs and cats, and for the CUB200 experiment, we removed nine clusters that contained images of birds. The remaining clusters are used to train experts and in turn recommend data. 
From Table~\ref{tab:ablate}, we can see that dropping experts trained on classes of images in the target set did not have significant effect on performance. Hence, SNDS can be used in the real world without frequent updates to the public dataset once it is deployed. As a side note, we use sampling without replacement over the entire source dataset, similar to that used in NDS, for SNDS results reported in this experiment.

% \begin{wrapfigure}{R}{0.53\linewidth}
%     \vspace{-2mm}
%     \centering
%     \begin{subfigure}[b]{0.492\linewidth}
%     \centering
%     \includegraphics[trim=5 0 15 0,clip,width=\linewidth]{"Computational Cost Per Query with Growing Server (1)".pdf}
%     \end{subfigure}
%     \hfill
%      \begin{subfigure}[b]{0.492\linewidth}
%     \centering
%     \includegraphics[trim=5 0 15 0,clip,width=\linewidth]{"Total Computational Cost with Sources and Queries (1)".pdf}
%     \end{subfigure}
%     \vspace{-6mm}
%     \caption{\small Comparison of simulated computational cost. }
%     %Left: the cost of answering one query plotted over number of data sources index by the server. Right: total %computational cost incurred over time with number of sources and number of queries both growing over time.}
%     \label{fig:sim}
%     \vspace{-12pt}
% \end{wrapfigure}
%\begin{wrapfigure}{R}{0.59\linewidth}
    \vspace{-2mm}
    \begin{figure}[t!]
    \centering
    \begin{subfigure}[b]{0.43\linewidth}
    \centering
    \includegraphics[trim=5 0 15 0,clip,width=\linewidth]{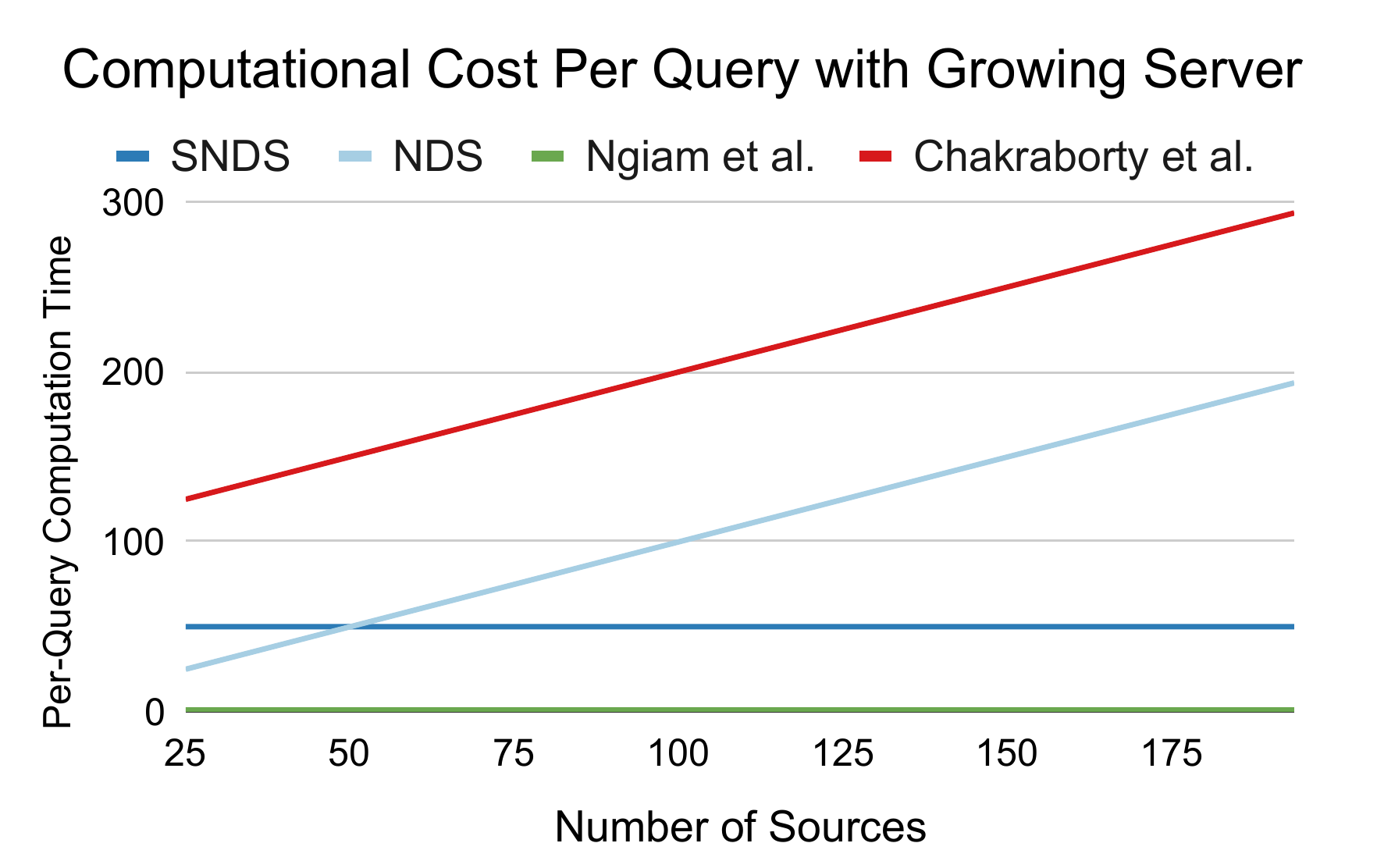}
    \end{subfigure}
    \hfill
     \begin{subfigure}[b]{0.43\linewidth}
    \centering
    \includegraphics[trim=5 0 15 0,clip,width=\linewidth]{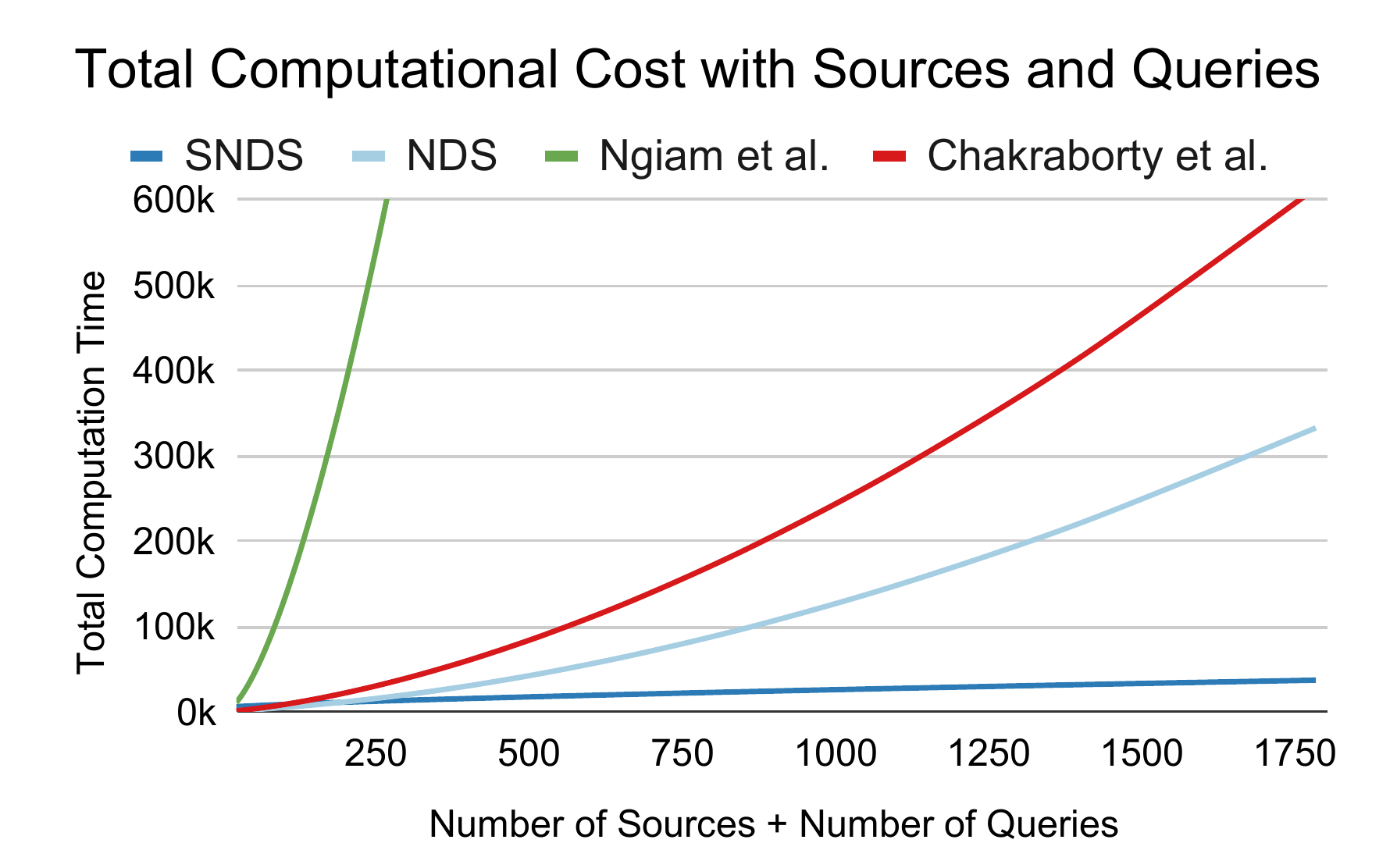}
    \end{subfigure}
    \vspace{-2mm}
    \caption{\small Comparison of simulated computational cost. }
    %Left: the cost of answering one query plotted over number of data sources index by the server. Right: total %computational cost incurred over time with number of sources and number of queries both growing over time.}
    \label{fig:sim}
    \vspace{2mm}
    \end{figure}
%\end{wrapfigure}

\vspace{-0mm}
\paragraph{Simulating Computational Cost} We simulate the computational cost of SNDS and compare it to that of \citet{yan2020neural}, \citet{ngiam2018domain}, and \citet{cui2018large} in Fig.~\ref{fig:sim}. We assume that the number of sources and the number of queries both grow over time, while the number of images per source stays constant. Detailed modeling assumptions can be found in the supplementary material. 
For the data consumer, SNDS has a flat per-query computational cost that does not grow with additional sources. Over time lifetime of a server, SNDS's total computational cost (incurred by all parties) is also the lowest among the methods compared.

%In the experiment described in Table~\ref{tab:snds1}, we used 50 experts which is equal to the number of sources. While we expect the number of sources to grow beyond the number of experts in a data search engine, the computational benefits of SNDS are not realized in that experiment. 
%Here, we study whether SNDS can recommend useful data with a much smaller set of experts.
 %in some downstream tasks such as CUB200, but it had a very small impact in most tasks. %In conjunction with Sec.~\ref{sec:missing_data}, we find that fine-grained splitting of the public dataset is more important than the content coverge.

\vspace{-0mm}
\subsubsection{Scaling beyond the Natural Image Setting} \label{sec:beyondNatural}
% \paragraph{Scaling beyond the natural image setting}
\vspace{-0mm}
The computational benefits of SNDS stem from using a fixed set of experts for performing data recommendation. For the system to scale, these experts must be able to provide clients with recommendations in domains beyond the ones that they has been trained on. 
% We have seen that SNDS provides meaningful recommendations for natural image datasets while using OpenImages' clusters as the source datasets. 
To test this, we explore the performance of the server, both qualitatively and quantitatively, when recommending data for domains that are highly dissimilar to natural images and find that SNDS is able to provide useful recommendations using experts that have been trained on ImageNet.
Note that in these experiments the public experts are kept the same, i.e. experts trained on ImageNet.
% \begin{wrapfigure}{L}{0.5\textwidth}
%     % \centering
    
%     \caption{Examples from the datasets used in our out-of-domain experiments}
%     \label{fig:domain_confusion}
%     % \vspace{-9pt}
% \end{wrapfigure}

\vspace{-0mm}
\paragraph{Medical data}

%\showthe\textwidth 

%\begin{wrapfigure}{R}{0.3\textwidth}
% \begin{figure*}[t!]
%     \vspace{-2mm}
%     \centering
%     % \begin{subfigure}[t]{0.5\textwidth}
%     % \centering
%     % \includegraphics[width=0.43\textwidth]{recommended_with_label.png}
%     \includegraphics[width=0.43\textwidth]{recommended_with_label.png}
%     \hfill
%     \includegraphics[width=0.275\textwidth]{sampling_output_nih.pdf}
%     % \end{subfigure}
%     \hfill
%     %  \begin{subfigure}[t]{0.5\textwidth}
%     % \centering
%     \includegraphics[width=0.275\textwidth]{sampling_output_aptos.pdf}
%     % \end{subfigure}
%     \vspace{-2mm}
%     \caption{Recommended data and output probabilities for satellite and x-ray data}
%     \label{fig:sampling-distributions}
%     \vspace{-20pt}
% %\end{wrapfigure}
% \end{figure*}

\begin{figure*}[t!]
    \vspace{-2mm}
    \centering
    % \begin{subfigure}[t]{0.5\textwidth}
    % \centering
    % \includegraphics[width=0.43\textwidth]{recommended_with_label.png}
    \includegraphics[width=0.92\textwidth,trim=150 360 0 0,clip]{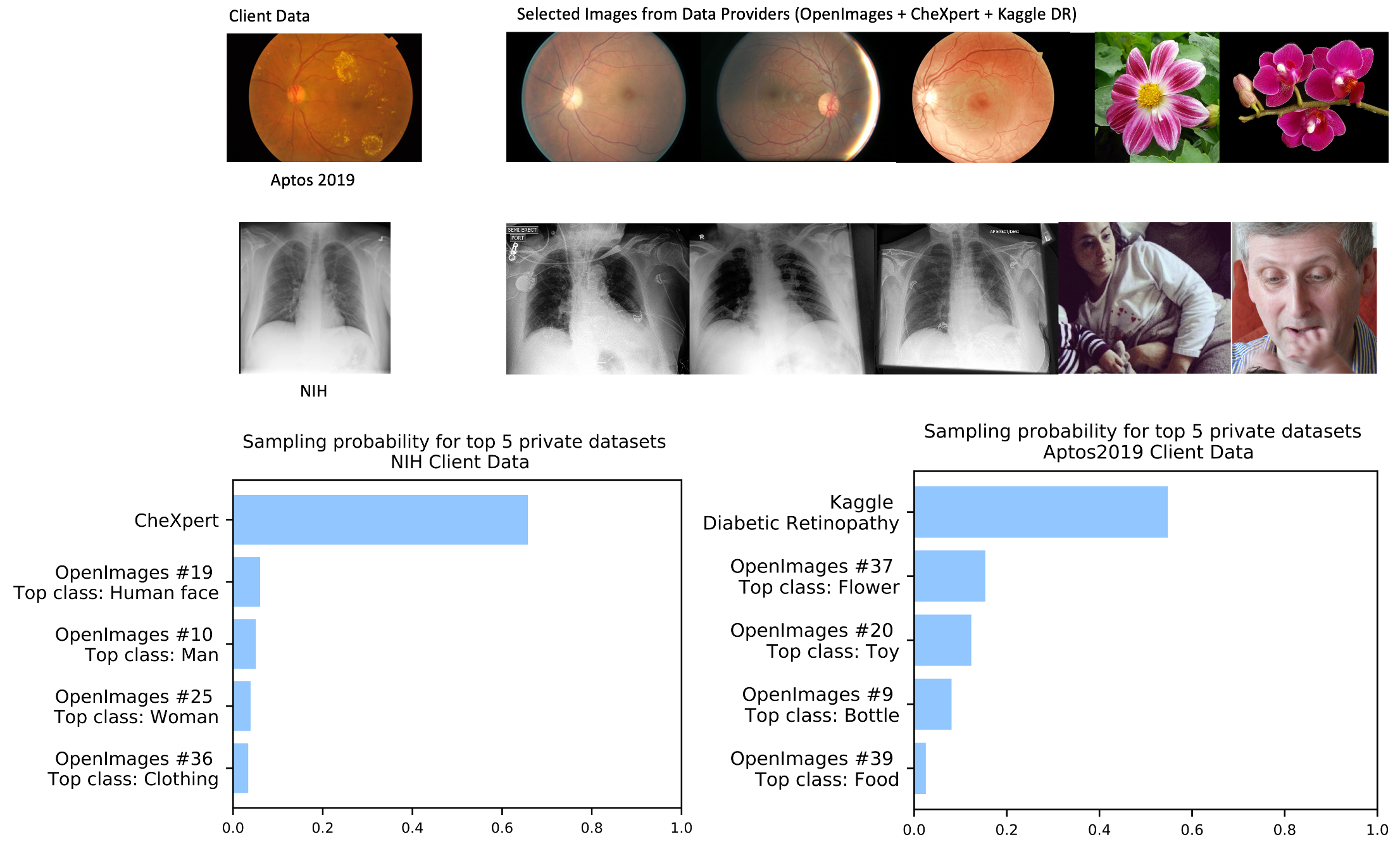}\\
      \includegraphics[width=0.82\textwidth,trim=0 0 0 330,clip]{figure_6_combined_two_rows.png}
    % \hfill
    % \includegraphics[width=0.275\textwidth]{sampling_output_nih.pdf}
    % % \end{subfigure}
    % \hfill
    % %  \begin{subfigure}[t]{0.5\textwidth}
    % % \centering
    % \includegraphics[width=0.275\textwidth]{sampling_output_aptos.pdf}
    % \end{subfigure}
    \vspace{-2mm}
    \caption{Recommended data and output probabilities for satellite and x-ray data}
    \label{fig:sampling-distributions}
    \vspace{-2mm}
%\end{wrapfigure}
\end{figure*}
% Medical imaging datasets differ substantially from natural image datasets, making them a challenging setting for transfer learning. Aside from differences in the image characteristics, these datasets typically have fewer classes, imbalanced data, higher resolution images and a focus on local texture rather than global information \cite{raghu2019transfusion}. Due to the smaller size of these datasets, transfer learning, especially with ImageNet, is commonly used in practice. However, \citet{raghu2019transfusion} find that this initialization gives little benefit to the downstream performance, and there have been many follow-up works exploring transfer learning with medical imaging, such as in \cite{Ke_2021, azizi2021big}.

Here, we focus on whether SNDS is able to recommend meaningful medical data with experts trained only on ImageNet. We add the chest x-ray dataset CheXpert \cite{irvin2019chexpert} and the diabetic retinopathy dataset provided by Kaggle \cite{kaggleDataset, cuadros2009eyepacs} to the original source datasets consisting of OpenImages clusters.
We examine the recommendations that the server provides when using different targets datasets of the same modalities as the source datasets: the NIH ChestX-ray14 dataset \cite{Wang2017}, and the Aptos 2019 diabetic retinopathy dataset \cite{aptos2019}. We do not perform any additional pre-processing to these datasets when incorporating them within SNDS.

In Fig.~\ref{fig:sampling-distributions}, we show the softmax scores of the top 5 recommended clusters for both the NIH ChestX-ray14 and Aptos2019 datasets. In both cases, the most relevant cluster in the source data corresponds to the image type in target data. Furthermore, SNDS does not conflate the recommendations between the chest x-ray and diabetic retinopathy, suggesting that the server is not grouping all dissimilar data together and maintains its ability to provide valuable recommendations in highly dissimilar data.

% We do not directly use this data for downstream evaluations, as transfer learning in the medical setting does not show consistent benefits (\cite{raghu2019transfusion}), and it remains an active area of research. These datasets are 
% \sasha{Furthermore, these datasets are multi-label and highly imbalanced, which makes it challenging to compare among baselines, which }

In practice, the best method for pre-training with medical images remains an open question and may not provide consistent benefits \cite{raghu2019transfusion}, though recent works in self-supervised learning have shown improvements when using a multi-step training process \cite{azizi2021big}. In our experiments, a naive supervised learning approach did not provide significant transfer benefits, even when directly transferring from one x-ray dataset to another (the oracle recommendation scenario). Still, the relevant recommendations provided by our server are directly usable by the consumer for improving performance using the latest techniques in this area as they emerge. 

\vspace{-2mm}
\paragraph{Sketch and satellite data evaluation}
\begin{wrapfigure}{R}{0.62\textwidth}
    \vspace{-15pt}
    \centering
    \small
    \captionof{table}{\small Task Performance for Satellite \& Sketch Data}
    \vspace{-2mm}
    \resizebox{0.615\textwidth}{!}{
    \addtolength{\tabcolsep}{-4pt}
    \begin{tabular}{l|c|c|c}
    \toprule
    \multirow{2}{*}{Selection Method} & \multirow{2}{*}{$\%$ Images} & \multicolumn{2}{c}{Target Dataset} \\
    & & Sketch-Mini & NWPU-RESISC45-Mini \\
    \midrule
    Rnd. Sample (+ sketch/sat)& $0.5\%$ (73K)& 11.79 & 52.40 \\
    \midrule
    SNDS (OpenImages only) & $0.5\%$ (73K)& 13.60 & 53.60 \\
    SNDS (+ sketch/sat data) & $0.5\%$ (73K)& 22.50 & 54.80 \\
    \midrule
    \bottomrule
    \end{tabular}%
    }
    \label{tab:snds-sketch-sat}
    \vspace{-6pt}
\end{wrapfigure}
Due to the challenges associated with transfer learning in the medical setting, we evaluate the benefits of transfer on more balanced datasets consisting of satellite and sketch datasets. We add the QuickDraw dataset split from \cite{peng2019moment} and the PatternNet dataset \cite{Zhou_2018} to the source sets of our server. As the client, we evaluate the performance of our method when using 1k subsets of the Sketch \cite{eitz2012hdhso} and NWPU-RESISC45 datasets \cite{Cheng_2017}, which we denote as Mini versions of those datasets. We use these subsets rather than the full datasets to better observe the effects of pretraining, as was done in NDS. To perform pretraining with the recommended data, we mix the various datasets together and perform standard supervised training with cross-entropy loss, predicting the class of the image from the union of all the dataset classes. 

Table~\ref{tab:snds-sketch-sat} provides performance for the satellite and sketch datasets. In both domains, SNDS shows performance benefits over a random sample of data from the server. SNDS even shows benefits over a random sample \textit{without} satellite and sketch dataset in the server, which suggests that our server is able to find natural images that help with downstream performance, even in dissimilar domains. The experts used for recommendation have been trained exclusively on ImageNet clusters, demonstrating that SNDS is truly able to scale and improve task performance in highly dissimilar domains. 

We see transfer benefits using our simple pre-training approach, which leads us to believe that consumers will be able to benefit even more from the data recommendations when using more complex approaches from multi-task learning. This highlights the strength and robustness of the SNDS' recommendations, which provides immediate benefits with a simple transfer algorithm. 

% Our quantiative results are shown 

% We wish to determine whether our approach truly scales to unseen domains outside of natural images, allowing data providers to upload arbitrary private datasets without changes to the public experts. To this end, we add several private datasets from the medical imaging and satellite imaging domain: CheXpert \cite{}, PadChest \cite{}, and EuroSAT \cite{} \textit{without} modifying the ImageNet experts. These were chosen due to their large size and ubiquity in those domains. Next, we test whether SNDS would be able to properly compute similarity and recommend relevant data distributions for client data consisting of chest x-ray images: the NIH dataset \cite{} and satellite images: the Sat-4 dataset \cite{}. 

\vspace{-0mm}
\paragraph{Limitations}\label{sec:limit} While SNDS can be applied to many task types, our experiments have focused on fine-grained classification tasks, allowing us to illustrate its behaviours across different data domains and in several important ablations. %Constraining the downstream task to fine-grained classification 
%This allowed us to focus our experiments on illustrating behaviors of SNDS through various important ablations. 
We also omit transfer experiments in the medical domain, as mentioned in Sec. \ref{sec:beyondNatural}. %on missing public data, changing the number of experts, and different data sampling strategies.
% We also show examples in several dissimilar domains, however we do not quantify the distance between 

%We do not perform transfer learning in the medical image domain, due to the challenges associated with transfer learning in that setting. 

% \begin{table}[h]
%     \centering
%     \small
%     \caption{Downstream task performance with pretraining data recommended by SNDS-10}
%     \begin{tabular}{lccc}
%     \toprule
%     \multirow{2}{*}{Selection Method} & \multirow{2}{*}{$\%$ Images} & \multicolumn{2}{c}{Target Dataset} \\
%     & & CUB200 & Oxford-IIIT \\
%     \midrule
%     Random Sample & $2\%$ \\
%     SNDS & $2\%$ \\
%     SNDS-10& $2\%$ \\
%     \midrule
%     Random Sample & $5\%$  \\
%     SNDS & $5\%$ \\
%     SNDS-10& $5\%$ \\
%     \bottomrule
%     \end{tabular}
%     \label{tab:10experts}
% \end{table}

\vspace{-0mm}
\section{Related Work}
\vspace{-0mm}

Our work is built upon the rich literature of transfer learning and domain adaptation \citep{zhuang2020comprehensive, pan2009survey,ben2007analysis,AcunaICML21_fDAL, mansour2009domain}. Of most resemblance to ours are the ``instance weighing" approaches ~\citep{sugiyama2008direct, sun2011two, belkin2006manifold, dai2007boosting, cui2018large, ngiam2018domain, achille2019task2vec, chakraborty2020efficient}. These methods compute the weight of instances or sets of instances based on some measure of usefulness to the target task. Our work differs from these methods in data separation and computational constraints: our problem setting requires a method that does can operate with the source and target dataset on separate devices, and the computational cost for indexing new source data must be low.

\citet{cui2018large} uses similarity between mean features of each class as extracted by a NN trained on source data
to match classes of the source to target dataset. 
\citet{ngiam2018domain} instead labels the target dataset with a NN trained on source data, and recommends source classes that appear most frequently as pseudo-labels.
%These two methods are both limited to recommending data from classification datasets.
These methods requires retraining of the NN whenever the source data is updated, thus they are computationally expensive in terms of total cost.

\citet{achille2019task2vec} find the Fisher information matrix for each source and target dataset to compute a similarity measure between sources and targets. This approach requires access to the consumer's labels and is challenging to scale beyond classification tasks, and thus not immediately applicable to our setting where we focus on being label-agnostic  to allow scalability to arbitrary tasks.

\citet{chakraborty2020efficient} proposed to train a NN on the target dataset, and compute similarity between the target dataset and each source in the feature space of this NN. 
Finally, \citet{yan2020neural} proposed NDS, reviewed in Sec.~\ref{sec:background}. These two methods have a per-query computational cost that grows with the number of sources. Furthermore, all existing methods require training of NNs on either sources or targets. This violates our privacy constraints (D1 and D2 in desiderata). SNDS protects the privacy of source and target data, and has a lower computational cost than existing methods. %In these methods, the computational cost of each query grows with the number of sources, whereas SNDS features a constant per-query cost.  %which trains NNs for a self-supervised task on partitions of the source dataset, and uses the performance of these NNs on the target as a marker of the target's similarity to each partition. \tianshi{contrast sentence for SNDS} We compare simulated computational cost of \citet{ngiam2018domain}, \citet{chakraborty2020efficient}, \citet{yan2020neural} and our proposed method in figure~\ref{fig:sim}, with details in Appendix~\ref{appendix:efficiency}.

\vspace{-0mm}
\section{Conclusion}
\vspace{-0mm}

We proposed SNDS, a scalable data recommendation method that selects sets of useful pretraining data for each given downstream task. 
In SNDS, sources and targets are represented by their degree of similarity to public datasets. %This similarity is computed by training experts on each public dataset and evaluating the experts on the sources and targets. 
%SNDS uses the similarity of each source and target dataset to a set of known public datasets to represent that dataset. 
%SNDS use the performance characteristics of an ensemble of experts on each dataset as a compact representation for that dataset. 
%This allows SNDS to index datasets by iterating through them once to obtain performance features, and then answer queries by matching in the feature space. 
The biggest advantage of SNDS over existing methods is computational efficiency: the computational cost for data consumers remains constant with the number of sources indexed by the server, and sources already indexed by SNDS do not need to be re-indexed when adding new sources. Experiments show that SNDS can work well even if there is data in the public datasets that is dissimilar to target, thereby allowing SNDS to generalize beyond what is contained in the public datasets. %We also show experimentally that SNDS can be applied to many downstream classification tasks. 
Lastly, we demonstrated the robustness of SNDS  by applying it to a variety of domains dissimilar to natural images and show transfer benefits in that setting.  

\vspace{-3mm}
\paragraph{Potential Societal Impact}\label{sec:impact} SNDS is a large-scale search engine for transfer learning data. Users can use SNDS to find relevant pretraining data for their dataset, which allows practitioners with smaller datasets to leverage the latest models designed for larger datasets, and to select pretraining data according to their computational needs. This is especially beneficial for consumers for whom collecting a large dataset is cost-prohibitive, such as in hospitals or non-profit organizations.
As a data recommendation search engine, our system is susceptible to misuses. % from both data providers and consumers. 
In a completely open system, data providers may upload inappropriate datasets or biased data to intentionally or unintentionally cause harm to the consumer. This can be mitigated by having a dataset approval process for SNDS and by including the latest tools to explore dataset biases as part of the server. %Consumers may also use SNDS for malicious applications, which is more difficult to detect, but may be mitigated by requiring approval for sensitive datasets in the server, such as medical images.

\paragraph{Acknowledgment} 
This work was partially supported by LG and NSERC. SF acknowledges the Canada CIFAR AI Chair award at the Vector Institute. TC acknowledges support from Vector Scholarship for Artificial Intelligence. DA acknowledges support from the Microsoft Ada Lovelace Fellowship.

\newpage
\bibliography{main}

\begin{thebibliography}{49}
\providecommand{\natexlab}[1]{#1}
\providecommand{\url}[1]{\texttt{#1}}
\expandafter\ifx\csname urlstyle\endcsname\relax
  \providecommand{\doi}[1]{doi: #1}\else
  \providecommand{\doi}{doi: \begingroup \urlstyle{rm}\Url}\fi

\bibitem[apt()]{aptos2019}
Aptos 2019 diabetic retinopathy dataset.
\newblock URL
  \url{https://www.kaggle.com/c/aptos2019-blindness-detection/data}.

\bibitem[kag(2015)]{kaggleDataset}
Diabetic retinopathy detection, 2015.
\newblock URL \url{https://www.kaggle.com/c/diabetic-retinopathy-detection}.

\bibitem[Achille et~al.(2019)Achille, Lam, Tewari, Ravichandran, Maji, Fowlkes,
  Soatto, and Perona]{achille2019task2vec}
A.~Achille, M.~Lam, R.~Tewari, A.~Ravichandran, S.~Maji, C.~Fowlkes, S.~Soatto,
  and P.~Perona.
\newblock Task2vec: Task embedding for meta-learning, 2019.

\bibitem[Acuna et~al.(2021)Acuna, Zhang, Law, and Fidler]{AcunaICML21_fDAL}
D.~Acuna, G.~Zhang, M.~Law, and S.~Fidler.
\newblock f-domain-adversarial learning: Theory and algorithms.
\newblock In \emph{Proceedings of the 38th International Conference on Machine
  Learning}, pages 66--75. PMLR, 2021.

\bibitem[AIMultiple(2020)]{aimultiple2020}
AIMultiple.
\newblock Data marketplaces: What, why, how, types, benefits, vendors, 2020.
\newblock URL \url{https://research.aimultiple.com/data-marketplace/#:~:text=A
  data marketplace is a upload data to the cloud.}

\bibitem[Azizi et~al.(2021)Azizi, Mustafa, Ryan, Beaver, Freyberg, Deaton, Loh,
  Karthikesalingam, Kornblith, Chen, Natarajan, and Norouzi]{azizi2021big}
S.~Azizi, B.~Mustafa, F.~Ryan, Z.~Beaver, J.~Freyberg, J.~Deaton, A.~Loh,
  A.~Karthikesalingam, S.~Kornblith, T.~Chen, V.~Natarajan, and M.~Norouzi.
\newblock Big self-supervised models advance medical image classification,
  2021.

\bibitem[Belkin et~al.(2006)Belkin, Niyogi, and Sindhwani]{belkin2006manifold}
M.~Belkin, P.~Niyogi, and V.~Sindhwani.
\newblock Manifold regularization: A geometric framework for learning from
  labeled and unlabeled examples.
\newblock \emph{Journal of machine learning research}, 7\penalty0
  (Nov):\penalty0 2399--2434, 2006.

\bibitem[Ben-David et~al.(2007)Ben-David, Blitzer, Crammer, and
  Pereira]{ben2007analysis}
S.~Ben-David, J.~Blitzer, K.~Crammer, and F.~Pereira.
\newblock Analysis of representations for domain adaptation.
\newblock In \emph{Advances in neural information processing systems}, pages
  137--144, 2007.

\bibitem[Bossard et~al.(2014)Bossard, Guillaumin, and
  Van~Gool]{bossard2014food}
L.~Bossard, M.~Guillaumin, and L.~Van~Gool.
\newblock Food-101--mining discriminative components with random forests.
\newblock In \emph{European conference on computer vision}, pages 446--461.
  Springer, 2014.

\bibitem[Chakraborty et~al.(2020)Chakraborty, Uzkent, Ayush, Tanmay, Sheehan,
  and Ermon]{chakraborty2020efficient}
S.~Chakraborty, B.~Uzkent, K.~Ayush, K.~Tanmay, E.~Sheehan, and S.~Ermon.
\newblock Efficient conditional pre-training for transfer learning.
\newblock \emph{arXiv preprint arXiv:2011.10231}, 2020.

\bibitem[Chen et~al.(2020)Chen, Jia, and Qi]{chen2020improved}
S.~Chen, R.~Jia, and G.-J. Qi.
\newblock Improved techniques for model inversion attack.
\newblock \emph{arXiv preprint arXiv:2010.04092}, 2020.

\bibitem[Cheng et~al.(2017)Cheng, Han, and Lu]{Cheng_2017}
G.~Cheng, J.~Han, and X.~Lu.
\newblock Remote sensing image scene classification: Benchmark and state of the
  art.
\newblock \emph{Proceedings of the IEEE}, 105\penalty0 (10):\penalty0
  1865–1883, Oct 2017.
\newblock ISSN 1558-2256.
\newblock \doi{10.1109/jproc.2017.2675998}.
\newblock URL \url{http://dx.doi.org/10.1109/JPROC.2017.2675998}.

\bibitem[Cimpoi et~al.(2014)Cimpoi, Maji, Kokkinos, Mohamed, , and
  Vedaldi]{cimpoi14describing}
M.~Cimpoi, S.~Maji, I.~Kokkinos, S.~Mohamed, , and A.~Vedaldi.
\newblock Describing textures in the wild.
\newblock In \emph{Proceedings of the {IEEE} Conf. on Computer Vision and
  Pattern Recognition ({CVPR})}, 2014.

\bibitem[Cuadros and Bresnick(2009)]{cuadros2009eyepacs}
J.~Cuadros and G.~Bresnick.
\newblock Eyepacs: an adaptable telemedicine system for diabetic retinopathy
  screening.
\newblock \emph{Journal of diabetes science and technology}, 3\penalty0
  (3):\penalty0 509--516, 2009.

\bibitem[Cui et~al.(2018)Cui, Song, Sun, Howard, and Belongie]{cui2018large}
Y.~Cui, Y.~Song, C.~Sun, A.~Howard, and S.~Belongie.
\newblock Large scale fine-grained categorization and domain-specific transfer
  learning.
\newblock In \emph{Proceedings of the IEEE conference on computer vision and
  pattern recognition}, pages 4109--4118, 2018.

\bibitem[Dai et~al.(2007)Dai, Yang, Xue, and Yu]{dai2007boosting}
W.~Dai, Q.~Yang, G.-R. Xue, and Y.~Yu.
\newblock Boosting for transfer learning.
\newblock In \emph{Proceedings of the 24th international conference on Machine
  learning}, pages 193--200, 2007.

\bibitem[Deng et~al.(2009)Deng, Dong, Socher, Li, Li, and
  Fei-Fei]{imagenet_cvpr09}
J.~Deng, W.~Dong, R.~Socher, L.-J. Li, K.~Li, and L.~Fei-Fei.
\newblock {ImageNet: A Large-Scale Hierarchical Image Database}.
\newblock In \emph{CVPR09}, 2009.

\bibitem[Eitz et~al.(2012)Eitz, Hays, and Alexa]{eitz2012hdhso}
M.~Eitz, J.~Hays, and M.~Alexa.
\newblock How do humans sketch objects?
\newblock \emph{ACM Trans. Graph. (Proc. SIGGRAPH)}, 31\penalty0 (4):\penalty0
  44:1--44:10, 2012.

\bibitem[Fredrikson et~al.(2015)Fredrikson, Jha, and
  Ristenpart]{fredrikson2015model}
M.~Fredrikson, S.~Jha, and T.~Ristenpart.
\newblock Model inversion attacks that exploit confidence information and basic
  countermeasures.
\newblock In \emph{Proceedings of the 22nd ACM SIGSAC Conference on Computer
  and Communications Security}, pages 1322--1333, 2015.

\bibitem[Fysical(2020)]{fysical2020}
Fysical.
\newblock The next data frontier isn't digital. it's fysical, 2020.
\newblock URL \url{https://fysical.org/}.

\bibitem[Ganin et~al.(2016)Ganin, Ustinova, Ajakan, Germain, Larochelle,
  Laviolette, Marchand, and Lempitsky]{ganin2016domain}
Y.~Ganin, E.~Ustinova, H.~Ajakan, P.~Germain, H.~Larochelle, F.~Laviolette,
  M.~Marchand, and V.~Lempitsky.
\newblock Domain-adversarial training of neural networks.
\newblock \emph{The journal of machine learning research}, 17\penalty0
  (1):\penalty0 2096--2030, 2016.

\bibitem[Goodfellow et~al.(2016)Goodfellow, Bengio, Courville, and
  Bengio]{goodfellow2016deep}
I.~Goodfellow, Y.~Bengio, A.~Courville, and Y.~Bengio.
\newblock \emph{Deep learning}, volume~1.
\newblock MIT Press, 2016.

\bibitem[Irvin et~al.(2019)Irvin, Rajpurkar, Ko, Yu, Ciurea-Ilcus, Chute,
  Marklund, Haghgoo, Ball, Shpanskaya, Seekins, Mong, Halabi, Sandberg, Jones,
  Larson, Langlotz, Patel, Lungren, and Ng]{irvin2019chexpert}
J.~Irvin, P.~Rajpurkar, M.~Ko, Y.~Yu, S.~Ciurea-Ilcus, C.~Chute, H.~Marklund,
  B.~Haghgoo, R.~Ball, K.~Shpanskaya, J.~Seekins, D.~A. Mong, S.~S. Halabi,
  J.~K. Sandberg, R.~Jones, D.~B. Larson, C.~P. Langlotz, B.~N. Patel, M.~P.
  Lungren, and A.~Y. Ng.
\newblock Chexpert: A large chest radiograph dataset with uncertainty labels
  and expert comparison, 2019.

\bibitem[Khosla et~al.(2011)Khosla, Jayadevaprakash, Yao, and
  Fei-Fei]{KhoslaYaoJayadevaprakashFeiFei_FGVC2011}
A.~Khosla, N.~Jayadevaprakash, B.~Yao, and L.~Fei-Fei.
\newblock Novel dataset for fine-grained image categorization.
\newblock In \emph{First Workshop on Fine-Grained Visual Categorization, IEEE
  Conference on Computer Vision and Pattern Recognition}, Colorado Springs, CO,
  June 2011.

\bibitem[Krause et~al.(2013)Krause, Stark, Deng, and
  Fei-Fei]{KrauseStarkDengFei-Fei_3DRR2013}
J.~Krause, M.~Stark, J.~Deng, and L.~Fei-Fei.
\newblock 3d object representations for fine-grained categorization.
\newblock In \emph{4th International IEEE Workshop on 3D Representation and
  Recognition (3dRR-13)}, Sydney, Australia, 2013.

\bibitem[Kuznetsova et~al.(2020)Kuznetsova, Rom, Alldrin, Uijlings, Krasin,
  Pont-Tuset, Kamali, Popov, Malloci, Kolesnikov, Duerig, and
  Ferrari]{OpenImages}
A.~Kuznetsova, H.~Rom, N.~Alldrin, J.~Uijlings, I.~Krasin, J.~Pont-Tuset,
  S.~Kamali, S.~Popov, M.~Malloci, A.~Kolesnikov, T.~Duerig, and V.~Ferrari.
\newblock The open images dataset v4: Unified image classification, object
  detection, and visual relationship detection at scale.
\newblock \emph{IJCV}, 2020.

\bibitem[Maji et~al.(2013)Maji, Kannala, Rahtu, Blaschko, and
  Vedaldi]{maji13fine-grained}
S.~Maji, J.~Kannala, E.~Rahtu, M.~Blaschko, and A.~Vedaldi.
\newblock Fine-grained visual classification of aircraft.
\newblock Technical report, 2013.

\bibitem[Mansour et~al.(2009)Mansour, Mohri, and
  Rostamizadeh]{mansour2009domain}
Y.~Mansour, M.~Mohri, and A.~Rostamizadeh.
\newblock Domain adaptation: Learning bounds and algorithms.
\newblock In \emph{Proceedings of The 22nd Annual Conference on Learning Theory
  (COLT 2009)}, Montr\'eal, Canada, 2009.
\newblock URL \url{http://www.cs.nyu.edu/~mohri/postscript/nadap.pdf}.

\bibitem[Neyshabur et~al.(2021)Neyshabur, Sedghi, and
  Zhang]{neyshabur2021transferred}
B.~Neyshabur, H.~Sedghi, and C.~Zhang.
\newblock What is being transferred in transfer learning?, 2021.

\bibitem[Ngiam et~al.(2018)Ngiam, Peng, Vasudevan, Kornblith, Le, and
  Pang]{ngiam2018domain}
J.~Ngiam, D.~Peng, V.~Vasudevan, S.~Kornblith, Q.~V. Le, and R.~Pang.
\newblock Domain adaptive transfer learning with specialist models.
\newblock \emph{arXiv preprint arXiv:1811.07056}, 2018.

\bibitem[Nilsback and Zisserman(2008)]{Nilsback08}
M.-E. Nilsback and A.~Zisserman.
\newblock Automated flower classification over a large number of classes.
\newblock In \emph{Indian Conference on Computer Vision, Graphics and Image
  Processing}, Dec 2008.

\bibitem[Pan and Yang(2009)]{pan2009survey}
S.~J. Pan and Q.~Yang.
\newblock A survey on transfer learning.
\newblock \emph{IEEE Transactions on knowledge and data engineering},
  22\penalty0 (10):\penalty0 1345--1359, 2009.

\bibitem[Parkhi et~al.(2012)Parkhi, Vedaldi, Zisserman, and Jawahar]{parkhi12a}
O.~M. Parkhi, A.~Vedaldi, A.~Zisserman, and C.~V. Jawahar.
\newblock Cats and dogs.
\newblock In \emph{IEEE Conference on Computer Vision and Pattern Recognition},
  2012.

\bibitem[Paszke et~al.(2019)Paszke, Gross, Massa, Lerer, Bradbury, Chanan,
  Killeen, Lin, Gimelshein, Antiga, Desmaison, Kopf, Yang, DeVito, Raison,
  Tejani, Chilamkurthy, Steiner, Fang, Bai, and Chintala]{NEURIPS2019_9015}
A.~Paszke, S.~Gross, F.~Massa, A.~Lerer, J.~Bradbury, G.~Chanan, T.~Killeen,
  Z.~Lin, N.~Gimelshein, L.~Antiga, A.~Desmaison, A.~Kopf, E.~Yang, Z.~DeVito,
  M.~Raison, A.~Tejani, S.~Chilamkurthy, B.~Steiner, L.~Fang, J.~Bai, and
  S.~Chintala.
\newblock Pytorch: An imperative style, high-performance deep learning library.
\newblock In H.~Wallach, H.~Larochelle, A.~Beygelzimer, F.~d\textquotesingle
  Alch\'{e}-Buc, E.~Fox, and R.~Garnett, editors, \emph{Advances in Neural
  Information Processing Systems 32}, pages 8024--8035. Curran Associates,
  Inc., 2019.
\newblock URL
  \url{http://papers.neurips.cc/paper/9015-pytorch-an-imperative-style-high-performance-deep-learning-library.pdf}.

\bibitem[Peng et~al.(2019)Peng, Bai, Xia, Huang, Saenko, and
  Wang]{peng2019moment}
X.~Peng, Q.~Bai, X.~Xia, Z.~Huang, K.~Saenko, and B.~Wang.
\newblock Moment matching for multi-source domain adaptation.
\newblock In \emph{Proceedings of the IEEE International Conference on Computer
  Vision}, pages 1406--1415, 2019.

\bibitem[Raghu et~al.(2019)Raghu, Zhang, Kleinberg, and
  Bengio]{raghu2019transfusion}
M.~Raghu, C.~Zhang, J.~Kleinberg, and S.~Bengio.
\newblock Transfusion: Understanding transfer learning for medical imaging.
\newblock \emph{arXiv preprint arXiv:1902.07208}, 2019.

\bibitem[Russakovsky et~al.(2015)Russakovsky, Deng, Su, Krause, Satheesh, Ma,
  Huang, Karpathy, Khosla, Bernstein, Berg, and Fei-Fei]{ILSVRC15}
O.~Russakovsky, J.~Deng, H.~Su, J.~Krause, S.~Satheesh, S.~Ma, Z.~Huang,
  A.~Karpathy, A.~Khosla, M.~Bernstein, A.~C. Berg, and L.~Fei-Fei.
\newblock {ImageNet Large Scale Visual Recognition Challenge}.
\newblock \emph{International Journal of Computer Vision (IJCV)}, 115\penalty0
  (3):\penalty0 211--252, 2015.
\newblock \doi{10.1007/s11263-015-0816-y}.

\bibitem[Snowflake(2020)]{Snowflake2020}
Snowflake.
\newblock Snowflake data marketplace, 2020.
\newblock URL \url{https://www.snowflake.com/data-marketplace/}.

\bibitem[Sugiyama et~al.(2008)Sugiyama, Suzuki, Nakajima, Kashima, von
  B{\"u}nau, and Kawanabe]{sugiyama2008direct}
M.~Sugiyama, T.~Suzuki, S.~Nakajima, H.~Kashima, P.~von B{\"u}nau, and
  M.~Kawanabe.
\newblock Direct importance estimation for covariate shift adaptation.
\newblock \emph{Annals of the Institute of Statistical Mathematics},
  60\penalty0 (4):\penalty0 699--746, 2008.

\bibitem[Sun et~al.(2011)Sun, Chattopadhyay, Panchanathan, and Ye]{sun2011two}
Q.~Sun, R.~Chattopadhyay, S.~Panchanathan, and J.~Ye.
\newblock A two-stage weighting framework for multi-source domain adaptation.
\newblock \emph{Advances in neural information processing systems},
  24:\penalty0 505--513, 2011.

\bibitem[VisualData(2020)]{visualio}
VisualData.
\newblock Visualdata - search engine for computer vision datasets, 2020.
\newblock URL \url{https://www.visualdata.io/discovery}.

\bibitem[Wang et~al.(2017)Wang, Peng, Lu, Lu, Bagheri, and Summers]{Wang2017}
X.~Wang, Y.~Peng, L.~Lu, Z.~Lu, M.~Bagheri, and R.~M. Summers.
\newblock Chestx-ray8: Hospital-scale chest x-ray database and benchmarks on
  weakly-supervised classification and localization of common thorax diseases.
\newblock \emph{2017 IEEE Conference on Computer Vision and Pattern Recognition
  (CVPR)}, Jul 2017.
\newblock \doi{10.1109/cvpr.2017.369}.
\newblock URL \url{http://dx.doi.org/10.1109/CVPR.2017.369}.

\bibitem[Welinder et~al.(2010)Welinder, Branson, Mita, Wah, Schroff, Belongie,
  and Perona]{WelinderEtal2010}
P.~Welinder, S.~Branson, T.~Mita, C.~Wah, F.~Schroff, S.~Belongie, and
  P.~Perona.
\newblock {Caltech-UCSD Birds 200}.
\newblock Technical Report CNS-TR-2010-001, California Institute of Technology,
  2010.

\bibitem[Xiao et~al.(2010)Xiao, Hays, Ehinger, Oliva, and
  Torralba]{xiao2010sun}
J.~Xiao, J.~Hays, K.~A. Ehinger, A.~Oliva, and A.~Torralba.
\newblock Sun database: Large-scale scene recognition from abbey to zoo.
\newblock In \emph{2010 IEEE computer society conference on computer vision and
  pattern recognition}, pages 3485--3492. IEEE, 2010.

\bibitem[Yan et~al.(2020)Yan, Acuna, and Fidler]{yan2020neural}
X.~Yan, D.~Acuna, and S.~Fidler.
\newblock Neural data server: A large-scale search engine for transfer learning
  data.
\newblock In \emph{Proceedings of the IEEE/CVF Conference on Computer Vision
  and Pattern Recognition}, pages 3893--3902, 2020.

\bibitem[Zamir et~al.(2018)Zamir, Sax, Shen, Guibas, Malik, and
  Savarese]{taskonomy2018}
A.~R. Zamir, A.~Sax, W.~B. Shen, L.~J. Guibas, J.~Malik, and S.~Savarese.
\newblock Taskonomy: Disentangling task transfer learning.
\newblock In \emph{IEEE Conference on Computer Vision and Pattern Recognition
  (CVPR)}. IEEE, 2018.

\bibitem[Zhang et~al.(2020)Zhang, Jia, Pei, Wang, Li, and
  Song]{zhang2020secret}
Y.~Zhang, R.~Jia, H.~Pei, W.~Wang, B.~Li, and D.~Song.
\newblock The secret revealer: Generative model-inversion attacks against deep
  neural networks.
\newblock In \emph{Proceedings of the IEEE/CVF Conference on Computer Vision
  and Pattern Recognition}, pages 253--261, 2020.

\bibitem[Zhou et~al.(2018)Zhou, Newsam, Li, and Shao]{Zhou_2018}
W.~Zhou, S.~Newsam, C.~Li, and Z.~Shao.
\newblock Patternnet: A benchmark dataset for performance evaluation of remote
  sensing image retrieval.
\newblock \emph{ISPRS Journal of Photogrammetry and Remote Sensing},
  145:\penalty0 197–209, Nov 2018.
\newblock ISSN 0924-2716.
\newblock \doi{10.1016/j.isprsjprs.2018.01.004}.
\newblock URL \url{http://dx.doi.org/10.1016/j.isprsjprs.2018.01.004}.

\bibitem[Zhuang et~al.(2020)Zhuang, Qi, Duan, Xi, Zhu, Zhu, Xiong, and
  He]{zhuang2020comprehensive}
F.~Zhuang, Z.~Qi, K.~Duan, D.~Xi, Y.~Zhu, H.~Zhu, H.~Xiong, and Q.~He.
\newblock A comprehensive survey on transfer learning.
\newblock \emph{Proceedings of the IEEE}, 2020.

\end{thebibliography}
\bibliographystyle{abbrvnat}
\newpage

\section*{Checklist}

%%% BEGIN INSTRUCTIONS %%%
% The checklist follows the references.  Please
% read the checklist guidelines carefully for information on how to answer these
% questions.  For each question, change the default \answerTODO{} to \answerYes{},
% \answerNo{}, or \answerNA{}.  You are strongly encouraged to include a {\bf
% justification to your answer}, either by referencing the appropriate section of
% your paper or providing a brief inline description.  For example:
% \begin{itemize}
%   \item Did you include the license to the code and datasets? \answerYes{See Section~\ref{gen_inst}.}
%   \item Did you include the license to the code and datasets? \answerNo{The code and the data are proprietary.}
%   \item Did you include the license to the code and datasets? \answerNA{}
% \end{itemize}
% Please do not modify the questions and only use the provided macros for your
% answers.  Note that the Checklist section does not count towards the page
% limit.  In your paper, please delete this instructions block and only keep the
% Checklist section heading above along with the questions/answers below.
%%% END INSTRUCTIONS %%%

\begin{enumerate}

\item For all authors...
\begin{enumerate}
  \item Do the main claims made in the abstract and introduction accurately reflect the paper's contributions and scope?
    \answerYes{We describe 3 main claims in the in the introduction (Section \ref{sec:intro}) in terms of the generalization and performance of the system, both in natural image and dissimilar image domains, which we demonstrate experimentally in Sections \ref{sec:mainExperiments} and \ref{sec:beyondNatural}. We also describe the scalability of the system in detail in terms of computational cost within \ref{sec:snds} and simulate the computational growth in Fig.~\ref{fig:sim}. }
  \item Did you describe the limitations of your work?
    \answerYes{See Section \ref{sec:limit}}
  \item Did you discuss any potential negative societal impacts of your work?
    \answerYes{See Section \ref{sec:impact}}
  \item Have you read the ethics review guidelines and ensured that your paper conforms to them?
    \answerYes{In addition to reading the guidelines, we discuss the broader implications of our work in Section \ref{sec:impact} as mentioned in the previous question.}
\end{enumerate}

\item If you are including theoretical results...
\begin{enumerate}
  \item Did you state the full set of assumptions of all theoretical results?
    \answerNA{}
	\item Did you include complete proofs of all theoretical results?
    \answerNA{}
\end{enumerate}

\item If you ran experiments...
\begin{enumerate}
  \item Did you include the code, data, and instructions needed to reproduce the main experimental results (either in the supplemental material or as a URL)?
    \answerNo{This is proprietary information that we will not be releasing at this time.}
  \item Did you specify all the training details (e.g., data splits, hyperparameters, how they were chosen)?
    \answerYes{We will be addressing these in the supplementary material.}
	\item Did you report error bars (e.g., with respect to the random seed after running experiments multiple times)?
    \answerNo{Due to the large-scale experiments required for transfer learning, we were unable to get these results for the paper, as this requires pre-training with a large set of data for a variety of datasets and finetuning on downstream tasks. This would require at least 3 times the current computational resources, making this computationally infeasible for the current submission.}
	\item Did you include the total amount of compute and the type of resources used (e.g., type of GPUs, internal cluster, or cloud provider)?
    \answerYes{See section \ref{sec:dataRecommendation} for the type of resource. Given the amount of experiments performed during the process (with different percentages of data across a wide variety of datasets), we were unable to properly quantify the total amount of resources used. }
\end{enumerate}

\item If you are using existing assets (e.g., code, data, models) or curating/releasing new assets...
\begin{enumerate}
  \item If your work uses existing assets, did you cite the creators?
    \answerYes{We mention the datasets in Section \ref{sec:datasets} and within the relevant parts of \ref{sec:beyondNatural}}
  \item Did you mention the license of the assets?
    \answerYes{This is included in the supplementary material.}
  \item Did you include any new assets either in the supplemental material or as a URL?
    \answerNA{}
  \item Did you discuss whether and how consent was obtained from people whose data you're using/curating?
    \answerYes{We did not curate any data ourselves, but used pre-existing datasets curated by others. Details are included in the supplementary material.}
  \item Did you discuss whether the data you are using/curating contains personally identifiable information or offensive content?
    \answerYes{This is included in the supplementary material.}
\end{enumerate}

\item If you used crowdsourcing or conducted research with human subjects...
\begin{enumerate}
  \item Did you include the full text of instructions given to participants and screenshots, if applicable?
    \answerNA{}{}
  \item Did you describe any potential participant risks, with links to Institutional Review Board (IRB) approvals, if applicable?
    \answerNA{}
  \item Did you include the estimated hourly wage paid to participants and the total amount spent on participant compensation?
    \answerNA{}
\end{enumerate}

\end{enumerate}

\onecolumn
\appendix
\section{Transfer Learning Details}
Experiments are performed in PyTorch, with licensing information here: \url{https://github.com/pytorch/pytorch/blob/master/LICENSE}.
Pretraining is performed using a ResNet18 as backbone on the selected data. We train for 40 epochs on the classification task with cross-entropy loss, and then finetune for 100 epochs on the target dataset with a new classification head. In Section \ref{sec:mainExperiments} the task is classification between the 601 classes in OpenImages, whereas in Sec\ref{sec:beyondNatural} we classify between all of the classes in the mixed data recommendations. We use performance on validation split of the target dataset for early stopping during finetuning. No early stopping is used during pretraining. 
SGD optimizer with a momentum $0.9$ is used for both pretraining and finetuning, with learning rates of $0.1$ and $0.01$ respectively. We decay the learning rate of the optimizer during finetuning by a factor of 10 every 30 epochs. Weight decay of $0.0001$ is applied during both steps. 
All images are normalized in each channel by Imagenet image normalization weights. For data augmentation during pretraining, we resize each image to a short-edge length of 256 and then randomly crop a 224x224 region from the image. Random horizontal flipping is also used. During finetuning, we additionally perform random rotation of up to 20 degrees. 

\section{Proof of triangle inequality}
Recall notation from the main text: $R_A(h, h')$ is the risk associated with predicting with hypothesis $h$ on data distribution $A$ when the true labeling function is $h'$. We define $disc_l(S,T)$ as:
\begin{equation}
    disc_l(S, T) = \sup_{(h,h') \in \mathcal{H}\times \mathcal{H}} \left|R^l_S(h',h) - R^l_T(h',h)  \right|
\end{equation}
Let $(h_{AC},h'_{AC})$ denote the pair of hypothesis that maximizes $\left|R^l_A(h',h) - R^l_C(h',h)  \right|$, then 
\begin{align}
    disc_l(A,C) &= \left|R^l_A(h_{AC},h'_{AC}) - R^l_C(h'_{AC},h_{AC})  \right| \\
        & \leq \left|R^l_A(h_{AC},h'_{AC}) - R^l_B(h'_{AC},h_{AC})  \right| \\
        &  \quad + \left|R^l_B(h_{AC},h'_{AC}) - R^l_C(h'_{AC},h_{AC})  \right| \\
        & \leq \sup_{(h,h') \in \mathcal{H}\times \mathcal{H}} \left|R^l_A(h,h') - R^l_B(h',h)  \right| \\
        & \quad + \sup_{(h,h') \in \mathcal{H}\times \mathcal{H}} \left|R^l_B(h',h) - R^l_C(h',h)  \right| \\
        &= disc_l(A,B) + disc_l(B,C).
\end{align}

\section{Description of Datasets}

The datasets used in our experiments are described in Table~\ref{tab:dataset-stats}. We report the appropriate metric (top-1 accuracy or mean-per-class accuracy) as outlined in the evaluation protocols for each dataset. The DTD and SUN397 dataset include multiple train/test splits and we report our results on the first split, as done in prior work. For the CheXpert dataset, we use the splits defined by \cite{neyshabur2021transferred}, which creates a larger validation set than the one in the original work. For the mini datasets, we use two separate 1k subsets for training and testing, chosen from the original training datasets while maintaining the class balance.

\begin{table*}[h!]
\footnotesize
\resizebox*{\linewidth}{!}{
\addtolength{\tabcolsep}{-1pt}
\begin{tabular}{c|c|c|c|c}
\toprule
Dataset & Images & Class Count & Evaluation Metric \\
\hline \hline
OpenImages~\cite{OpenImages} & 1,743,042(images)/ 14,610,229(boxes) & 601 & -  \\
Imagenet  ~\cite{imagenet_cvpr09} & 1,281,167  & 1000 & - \\
\hline 
FGVC-Aircraft  ~\cite{maji13fine-grained} & 3334(train) / 3333(val) & 102 & Mean Per-Class \\
Stanford Cars ~\cite{KrauseStarkDengFei-Fei_3DRR2013} & 8144(train) / 8041 (val) & 196  & Top-1 \\
CUB200 Birds ~\cite{WelinderEtal2010} & 5994(train) / 5794(val) & 200 & Top-1 \\
Stanford Dogs ~\cite{KhoslaYaoJayadevaprakashFeiFei_FGVC2011} & 12,000(train) / 8580(val) & 120 &  Top-1 \\
Describable Textures Dataset \cite{cimpoi14describing} & 1880(train) / 1880(val) & 47 &  Top-1\\
Flowers 102 ~\cite{Nilsback08} & 2040(train) / 6149(val) & 102 & Mean Per-Class \\
Food-101 ~\cite{bossard2014food} & 75,750(train) / 25,250(val) & 101 & Top-1 \\
Oxford-IIIT Pets ~\cite{parkhi12a} & 3680(train) / 3369(val) & 37  & Mean Per-Class \\
SUN397 ~\cite{xiao2010sun} &  76,128(train) / 10,875(val) & 397 & Top-1 \\
Kaggle Diabetic Retinopathy ~\cite{kaggleDataset, cuadros2009eyepacs} & 35,126 (train) / 53,576(test) & 5 & - \\
Aptos 2019 ~\cite{aptos2019} & 3662(train) / 1928 (test) & 5 & - \\
CheXpert ~\cite{irvin2019chexpert, neyshabur2021transferred} & 50,000 (train) / 50,000 (test) & 5 & - \\
NIH ChestX-ray14 dataset \cite{Wang2017} & 86,524 (train)/25,596(test) & 14 & - \\
PatternNet ~\cite{Zhou_2018} & 30,400(train) & 38 & Top-1 \\
NWPU-RESISC45 ~\cite{Cheng_2017} & 31,500(train) & 45 & Top-1 \\
QuickDraw ~\cite{peng2019moment}  & 120,750(train)/51,750(test) & 345 & Top-1 \\
Sketch ~\cite{eitz2012hdhso}  & 20,000 (train) & 250 & Top-1 \\
% CheXpert ~\cite{} & 50,000 (train) / 50,000 (test) & 14 & AUC \\
\bottomrule
\end{tabular}%
}
\vspace{-3mm}
\caption{\small An overview of the public, private and target datasets used in our experiments.}
\vspace{-2mm}
\label{tab:dataset-stats}
\end{table*}

OpenImages is released under the Apache License 2.0. Imagenet is released for research-purposes only, detailed at \url{https://image-net.org/download.php}. FCVC-Aircraft is released ``for non-commercial research purposes only". Stanfard Cars uses the same license as Imagenet. CUB200 Birds is provided by \cite{WelinderEtal2010} on their project website \url{http://www.vision.caltech.edu/visipedia/CUB-200.html}. Stanford Dogs dataset is derived from Imagenet, and made available at \url{http://vision.stanford.edu/aditya86/ImageNetDogs/}. Describable Textures Dataset is released for research purpose only. Flowers 102 is licensed under the GNU General Public License, version 2. Food-101 was curated by \cite{bossard2014food} under fair use of data from foodsplotting.com, which has ceased operation in 2018. Oxford-IIIT Pets is available under the Creative Commons Attribution-ShareAlike 4.0 International (CC-BY-SA 4.0). SUN397 is made available for academic research by authors of \cite{xiao2010sun} on \url{https://vision.princeton.edu/projects/2010/SUN/}. The Kaggle Diabetic Retinopathy dataset is released by \cite{cuadros2009eyepacs} for research purposes, as mentioned by the Kaggle organizers here: \url{https://www.kaggle.com/c/diabetic-retinopathy-detection/discussion/141968}. The Aptos2019 dataset is released by the Kaggle licensing agreement, which is permissable for ``non-commercial purposes only'' and ``academic research'' as seen in \url{https://www.kaggle.com/c/aptos2019-blindness-detection/discussion/107520}. The CheXpert dataset is available for ``personal, non-commercial research purposes only'', as seen on \url{https://stanfordmlgroup.github.io/competitions/chexpert/}. Note that this dataset has 5 classes used for evaluation purposes, but 14 are available in total. The NIH ChestX-ray14 dataset can be used in an ``unrestricted'' fashion according to the dataset creators as mentioned in \url{https://nihcc.app.box.com/v/ChestXray-NIHCC/file/249502714403}. 
The PatternNet dataset is provided for research purposes and must be cited by the users:  \url{https://sites.google.com/view/zhouwx/dataset}. The  NWPU-RESISC45 dataset is said to be ``publicly available for research purposes'', as per \cite{Cheng_2017}. We use the QuickDraw split available from \cite{peng2019moment}, who include a fair use notice allowing others to use it for academic research: \url{http://ai.bu.edu/M3SDA/}.The Sketch dataset is released under a Creative Commons Attribution 4.0 International License on their site: \url{http://cybertron.cg.tu-berlin.de/eitz/projects/classifysketch/}. 

The datasets we use were provided with consent by their original curators and do not contain personally identifiable information to our knowledge, and we do not introduce any data curation in our work.

\section{Computational Efficiency Simulation} \label{appendix:efficiency}
We compare the computational cost of different data recommendation methods through simulation. The goal of the simulation is to show how each method's cost grows with increasing number of sources indexed by the server and with increasing number of targets that the server need to produce recommendations for. 
\paragraph{Modelling assumptions} We assume that methods which requires neural network training will train for 100 epochs (e.g. SNDS trains experts on the public dataset, NDS trains experts on sources), each source and target have a size of 10000 units, and passing a unit through a neural network has an ``iteration time" of $10^{-4}$ seconds (passing one network over one source has a cost of 1). We make the simplifying assumption that both the inference time and training step time can be approximated by ``iter time" for ease of comparison.
We also assume that the time for matching two representations (e.g. computing a cosine distance in SNDS or \cite{ngiam2018domain}) is $1e-4$.
For SNDS, we assume that there are 50 splits of public data ($K=50$), each split also has 10000 units. For \cite{ngiam2018domain}, we assume that instead of averaging over source labels, they instead train a domain classifier that distinguishes between sources, such that the mode of the prediction of this classifier on the target dataset indicates the most similar source set. 

We consider the computational costs with $M$ data providers, target datasets of size $n$ and source datasets of size $m$. For SNDS, we assume that we have $K$ public splits of data and corresponding experts, as mentioned previously.  

For computational cost per query, we use the following formulas: 
\begin{itemize}
    \item \cite{ngiam2018domain}: $n\times \text{iter time} + M \times \text{match time}$ 
\item \cite{chakraborty2020efficient}: $n\times \text{iter time} \times \text{Num epoch} + M \times m \times \text{iter time} + M \times \text{match time}$
\item \cite{yan2020neural}: $M \times n \times \text{iter time} + M \times \text{match time}$ 
\item SNDS: $K \times n \times \text{iter time} + K \times M \times \text{match time}$ 
\end{itemize}

Note that for each data recommendation method, the cost of the term containing $n \times \text{iter time}$ dominates the term containing $M \times \text{match time}$. This is because $n \times \text{iter time}$ corresponds to a eval or train step with a neural network over the target dataset, while $M \times \text{match time}$ corresponds to the time to search among the source providers to convert a computed metric into a sampling probability. This match step is performed with very low dimensional vectors compared to the size of the target dataset (at most K dimensions in the case of SNDS). Hence, compared to \cite{yan2020neural}, SNDS' performance benefits stem from reducing the number of forward passes that the consumer is required to compute, with the match time term being negligible. This is seen graphically in Figure ~\ref{fig:sim}. 

For total computational cost, we assume that the number of queries grows $10\%$ faster than the number of sources grows. Our simulation starts at 10 sources and 10 queries, and grows to 320 sources and 1462 queries. 
%For methods that require retraining when sources are added, we assume a constant update interval where an update is performed when sources have grow by $20\%$; more frequent updates would be more computationally expensive. 
The total computational cost is approximated by integrating the per-query cost over the growth in queries and total indexing cost of $M$ sources, and then summed together. 

For the total indexing cost, we use the following formulas:
\begin{itemize}
    \item \cite{ngiam2018domain}: $m \times \text{iter time} \times \text{Num epoch} \times{M \times (M-1)}$
 \item \cite{yan2020neural}: $M \times m \times \text{iter time} \times \text{Num epoch}$
\item SNDS: $ M \times m \times \text{iter time} \times K$
\item \cite{chakraborty2020efficient} does not have a separate indexing cost since it trains a network for each query and then performs indexing; all computational costs are captured in the per-query cost. 
\end{itemize}

\section{Softmax Temperature Optimization}

Once we have computed similarity scores $\mathbf{z}$ for each data source, we obtain the source weights with a softmax to find the source weights $\mathbf{w} \in \mathbb{R}^M$. However, we empirically find that the temperature $\tau$ is an important hyperparameter to set within the softmax equation. Mathematically, as $\tau \to \infty$, the softmax output resembles a uniform distribution, where each data source has an equal probability. As $\tau \to 0^{+}$, the softmax output approaches an $\argmax$ output, assigning 1 to the closest data source and 0 probability to the others. Hence, our procedure to determine a softmax temperature defines a trade-off between sampling uniformly and a greedy $\argmax$ output.

To make this trade-off automatically, we constrain the output distribution of weights $\mathbf{w}$ to have a specific entropy value $H(\mathbf{w})$ \textit{across} datasets. Entropy allows us to measure how ``closely" the weights $\mathbf{w}$ resemble a uniform distribution, and by setting a target entropy that the weights must satisfy, the output distribution is forced to have the same balance between a uniform and greedy sampling across all datasets.  This approach outperforms a fixed  $\tau$ in practice.  We adjust $\tau$ with gradient descent, which stably converges within dozens of iterations as this is a convex problem. In all  experiments, we set the target entropy value to $1.5$ --  the first value we tried.

% We normalize $\mathbf{z}$  via softmax with temperature $\tau$ to obtain source weights $\mathbf{w} \in \mathbb{R}^M$. The temperature parameter determines the trade-off between sampling greedily from the most relevant clusters and sampling uniformly from all clusters. Rather than assuming a constant value for $\tau$, we find the $\tau$ that satisfies a target entropy value for $\mathbf{w}$. Intuitively, this allows us to maintain the same trade-off between greedy vs. uniform sampling across different consumer datasets, and this approach outperforms a fixed  $\tau$ in practice. We adjust $\tau$ with gradient descent, which stably converges within dozens of iterations. In all  experiments, we set the target entropy value to $1.5$ --  the first value we tried.

\section{Additional Experiments}

\begin{table*}[h]
    \centering
    \small
    \caption{Downstream task performance with different sampling strategies}
    \resizebox{\textwidth}{!}{\begin{tabular}{lccccccccccc}
    \toprule
    \multirow{2}{*}{Selection Method} & \multirow{2}{*}{$\%$ Images} & \multicolumn{9}{c}{Target Dataset} & \multirow{2}{*}{Average}\\
    & & Aircraft & Cars & CUB200 & Stanford Dogs & DTD & Flowers102 & Food 101 & Pets & SUN397 & \\
    \midrule
    Random Sample         & $2\%$ & 54.66  & 46.77 & 41.49 & 52.93 & 52.72 & 74.03 & 72.34 & 67.02 & 39.79  & 55.75 \\
    SNDS                  & $2\%$ & 55.91  & 47.37 & 50.59 & 55.07 & 56.28 & 81.66 & 73.73 & 71.06 & 40.04  & 59.07 \\
    SNDS Greedy Selection & $2\%$ & 53.46  & 43.86 & 50.54 & 55.12 & 54.36 & 79.53 & 72.16 & 69.91 & 38.97 & 57.55  \\
    %SNDS Semi-Stochastic  & $2\%$ & 59.58  & 47.38 & 54.85 & 58.65 & 55.59 & 84.76 & 74.35 & 73.86 & 41.16 & 61.13 \\
     Oracle (NDS) & $2\%$ &  57.95  & 52.95 & 49.91 & 54.64 & 55.32 & 78.32 & 73.35 & 70.71 & 42.27  & 59.53 \\
    \midrule
    Random Sample         & $5\%$ & 62.70 & 67.34 &  53.11 & 57.93 & 57.87 & 83.53 & 74.86 & 73.79 & 44.42  & 63.95\\
    SNDS                  & $5\%$ & 62.32 & 63.80 &  57.35 & 59.38 & 60.59 & 87.68 & 74.87 & 75.91 & 43.99  & 64.96 \\
    SNDS Greedy Selection & $5\%$ & 62.11 & 63.45 &	 56.58 & 57.02 & 60.64 & 86.91 & 75.26 & 74.72 & 43.34  & 64.45 \\
    %SNDS Semi-Stochastic  & $5\%$ & 65.17 & 67.80 &  57.80 & 59.11 & 62.50 & 89.96 & 76.88 & 76.32 & 43.82  & 66.60 \\ 
    Oracle (NDS) & $5\%$ & 63.76  & 66.56 &  59.04 & 58.99 & 60.27 & 85.10 & 75.58 & 75.18 & 45.98 & 65.61 \\
    \bottomrule
    \end{tabular}}
    \label{tab:greedy-selection}
\end{table*}

\paragraph{Ablating Sampling Strategy} \label{sec:sampling}
Here, we ablate the sampling strategy from the stochastic sampling strategy presented in the paper on the OpenImages source data. Specifically, one could also perform a greedy selection strategy similar to that used in \cite{cui2018large} for data recommendations. Once SNDS has returned the score for each source, we start from the source with the highest score and add data to the pretraining set until the budget has been exhausted. Table~\ref{tab:greedy-selection} compares the downstream accuracy of greedy selection to the stochastic sampling strategy. We focus this ablation on $2\%$ and $5\%$ budgets due to the long pre-training time when using $10\%$. We find that greedy selection tends to perform better than random sampling, but worse than stochastic selection. Upon further inspection, we find that performance on the pretraining task is significantly lower when using greedy selection than that achieved when using stochastic sampling ($25.02\%$ vs $29.44\%$ on $2\%$ samples, $31.61\%$ vs $34.40\%$ on $5\%$ samples).
This suggests that while the sources with the highest scores may be most similar to the target, they lack the class diversity needed to make a good pretraining set.

% \begin{subtable}[t]{0.3\linewidth}
%     \centering
%     \small
%     \resizebox{\linewidth}{!}{\begin{tabular}[t]{lcc}
%     \toprule
%     Selection Method &$\%$ Images &  Average Downstream Accuracy\\
%     \midrule
%     Random Sample         & $2\%$ (292K) & 55.75 \\
%     SNDS-50 Stochastic       & $2\%$ (292K) & 59.07 \\
%     SNDS-50 Greedy Selection & $2\%$ (292K) & 57.55  \\
%     \midrule
%     Random Sample         & $5\%$ (730K) & 63.95\\
%     SNDS-50 Stochastic       & $5\%$ (730K) & 64.96 \\
%     SNDS-50 Greedy Selection & $5\%$ (730K) & 64.45 \\
%     \bottomrule
%     \end{tabular}}
%     \caption{Different sampling strategies}
%     \label{tab:greedy-selection-small}
% \end{subtable}%
% \hfill

\begin{table*}[h]
    \centering
    \small
    \caption{Downstream task performance with pretraining data recommended by SNDS-10}
    \resizebox{\textwidth}{!}{\begin{tabular}{lccccccccccc}
    \toprule
    \multirow{2}{*}{Selection Method} & \multirow{2}{*}{$\%$ Images} & \multicolumn{9}{c}{Target Dataset} & \multirow{2}{*}{Average}\\
    & & Aircraft & Cars & CUB200 & Stanford Dogs & DTD & Flowers102 & Food 101 & Pets & SUN397 & \\
    \midrule
    Random Sample & $2\%$ & 54.66  & 46.77 & 41.49 & 52.93 & 52.72 & 74.03 & 72.34 & 67.02 & 39.79  & 55.75 \\
    SNDS          & $2\%$ & 55.91  & 47.37 & 50.59 & 55.07 & 56.28 & 81.66 & 73.73 & 71.06 & 40.04  & 59.07 \\
    SNDS-10       & $2\%$ & 57.55  & 50.86 & 43.92 & 53.21 & 55.00 & 81.64 & 71.97 & 69.92 & 40.82  & 58.32 \\
    \midrule
    Random Sample & $5\%$ & 62.70 & 67.34 &  53.11 & 57.93 & 57.87 & 83.53 & 74.86 & 73.79 & 44.42  & 63.95\\
    SNDS          & $5\%$ & 62.32 & 63.80 &  57.35 & 59.38 & 60.59 & 87.68 & 74.87 & 75.91 & 43.99  & 64.96 \\
    SNDS-10       & $5\%$ & 62.19 & 64.03 &  53.09 & 57.91 & 59.68 & 89.29 & 75.91 & 74.65 & 43.28  & 64.00\\
    \bottomrule
    \end{tabular}}
    \label{tab:snds10experts}
\end{table*}

\paragraph{Ablating Number of Experts}
In this ablation we study the effect of reducing the number of experts.
We use the same public dataset (ImageNet) as before, but split it 10 ways ($K=10$) to train 10 experts. We call this smaller dataserver SNDS-10. We proceed to evaluate SNDS-10 on the same target tasks as used in Table~\ref{tab:snds1}. Results are reported in Table~\ref{tab:snds10experts}. We find that using fewer experts decreases the performance of SNDS. Our hypothesis is that this is caused by a lack of signal for SNDS to characterize datasets. Specifically, datasets are represented in SNDS by the performance of experts, which is a $K$ dimensional vector with range $[0,1]$. The capacity of this representation space (i.e. covering number) has an exponential relationship with $K$. When $K$ is reduced from 50 to 10, datasets that had distinct representations are now mapped to similar coordinates, thereby decreasing SNDS's ability to identify useful sources.

\begin{table*}[h]
    \centering
    \small
    \caption{Top-1 Accuracy on the pre-training task. }
    \resizebox{\textwidth}{!}{\begin{tabular}{lccccccccccc}
    \toprule
    \multirow{2}{*}{Selection Method} & \multirow{2}{*}{$\%$ Images} & \multicolumn{9}{c}{Target Dataset} & \multirow{2}{*}{\shortstack{Average}}\\
    & & Aircraft & Cars & CUB200 & Stanford Dogs & DTD & Flowers102 & Food 101 & Pets & SUN397 & \\
    \midrule
    Random Sample & $2\%$ & \multicolumn{10}{c}{34.78} \\
    SNDS &                  $2\%$ & 27.98 & 29.99 & 30.84 & 30.77 & 28.86 & 26.22 & 29.97 & 30.85 & 29.44 & 29.44 \\
    SNDS Greedy Selection & $2\%$ & 22.21 & 27.07 & 26.12 & 29.48 & 26.54 & 15.81 & 23.61 & 27.91 & 26.41 & 25.02 \\
    \midrule
    Random Sample & $5\%$ & \multicolumn{10}{c}{39.53} \\
    SNDS                  & $5\%$ & 32.16 & 33.14 & 36.37 & 35.08 & 34.34 & 34.68 & 35.16 & 34.83 & 33.88 & 34.40 \\
    SNDS Greedy Selection & $5\%$ & 29.76 & 32.40 & 30.44 & 31.60 & 34.83 & 29.01 & 30.93 & 31.75 & 33.76 & 31.61 \\
    \bottomrule
    \end{tabular}}
    \label{tab:pretrain}
\end{table*}

\paragraph{Pretraining task accuracy} We present the accuracy achieved on the 601-way classification pretraining task with OpenImages sources in Table ~\ref{tab:pretrain}. For the ``Random Sample" baseline, all downstream tasks share the same pretrained network, hence only one performance number is reported.

\paragraph{Repeat runs with confidence intervals}
We perform repeat trials for the first three target datasets of Table~\ref{tab:snds1} for SNDS and NDS to obtain confidence intervals. In Table~\ref{tab:repeat}, we show the average and standard deviation over 3 runs with different seeds for SNDS and NDS. This corresponds to 3 different samples of data for each reported average.

\begin{table*}[h]
    \centering
    \small
    \caption{Downstream task performance with 3 repeated runs. }
    \begin{tabular}{lcccc}
    \toprule
    \multirow{2}{*}{Selection Method} & \multirow{2}{*}{$\%$ Images} & \multicolumn{3}{c}{Target Dataset} \\
    & & CUB200 & Flowers102 & Pets \\
    \midrule
    SNDS & $2\%$ & $50.24\pm 0.88$ & $82.66\pm 0.79$ & $70.94 \pm 1.17$ \\
    NDS & $2\%$ & $49.94 \pm 0.28$ & $79.67 \pm 0.80$ & $70.85 \pm 0.57$\\
    \midrule
    SNDS & $5\%$ & $57.67 \pm 1.05$ & $87.02 \pm 0.29$ & $75.42 \pm 0.74$ \\
    NDS & $5\%$ & $58.33 \pm 0.61$ & $86.15 \pm 0.13$ & $75.36\pm 0.57$\\
    \midrule
    SNDS & $10\%$ & $60.81 \pm 0.61$ & $89.94 \pm 0.52$ & $77.09 \pm 0.17$ \\
    NDS & $10\%$ &  $60.23\pm 1.59$ & $88.86 \pm 1.77$ & $77.35 \pm 0.96$ \\
    \bottomrule
    \end{tabular}
    \label{tab:repeat}
\end{table*}
% \begin{subtable}[t]{0.5\linewidth}
%     \centering
%     \small
%     \resizebox{0.9\linewidth}{!}{\begin{tabular}[t]{lcc}
%     \toprule
%     Selection Method &$\%$ Images &  Average Downstream Accuracy\\
%     \midrule
%     SNDS-10       & $2\%$ (292K) & 58.32 \\
%     SNDS-50& $2\%$ (292K) & 59.07 \\
%     \midrule
%     SNDS-10       & $5\%$ (730K) & 64.00 \\
%     SNDS-50 & $5\%$ (730K) &  64.96 \\
    
%     \bottomrule
%     \end{tabular}}
%     \caption{Downstream task performance with 10 experts}
%     \label{tab:snds10-small}
% \end{subtable}%

% \begin{table*}[h]
%     \centering
%     \small
%     \caption{Downstream task performance with pretraining data recommended by SNDS.}
%     \resizebox{\textwidth}{!}{\begin{tabular}{lccccccccccc}
%     \toprule
%     \multirow{2}{*}{Selection Method} & \multirow{2}{*}{$\%$ Images} & \multicolumn{9}{c}{Target Dataset} & \multirow{2}{*}{\shortstack{Average}}\\
%     & & Aircraft & Cars & CUB200 & Stanford Dogs & DTD & Flowers102 & Food 101 & Pets & SUN397 & \\
%     \midrule
%     Random Sample & $10\%$ &  \\
%     SNDS &          $10\%$ &  \\
%     Oracle (NDS) & $10\%$ &   \\
%     \bottomrule
%     \end{tabular}}
%     \label{tab:snds10pct}
% \end{table*}

\clearpage
\section{Additional Qualitative Results}
\begin{figure}[h]
    \vspace{-6mm}
    \centering
    \includegraphics[width=0.97\textwidth]{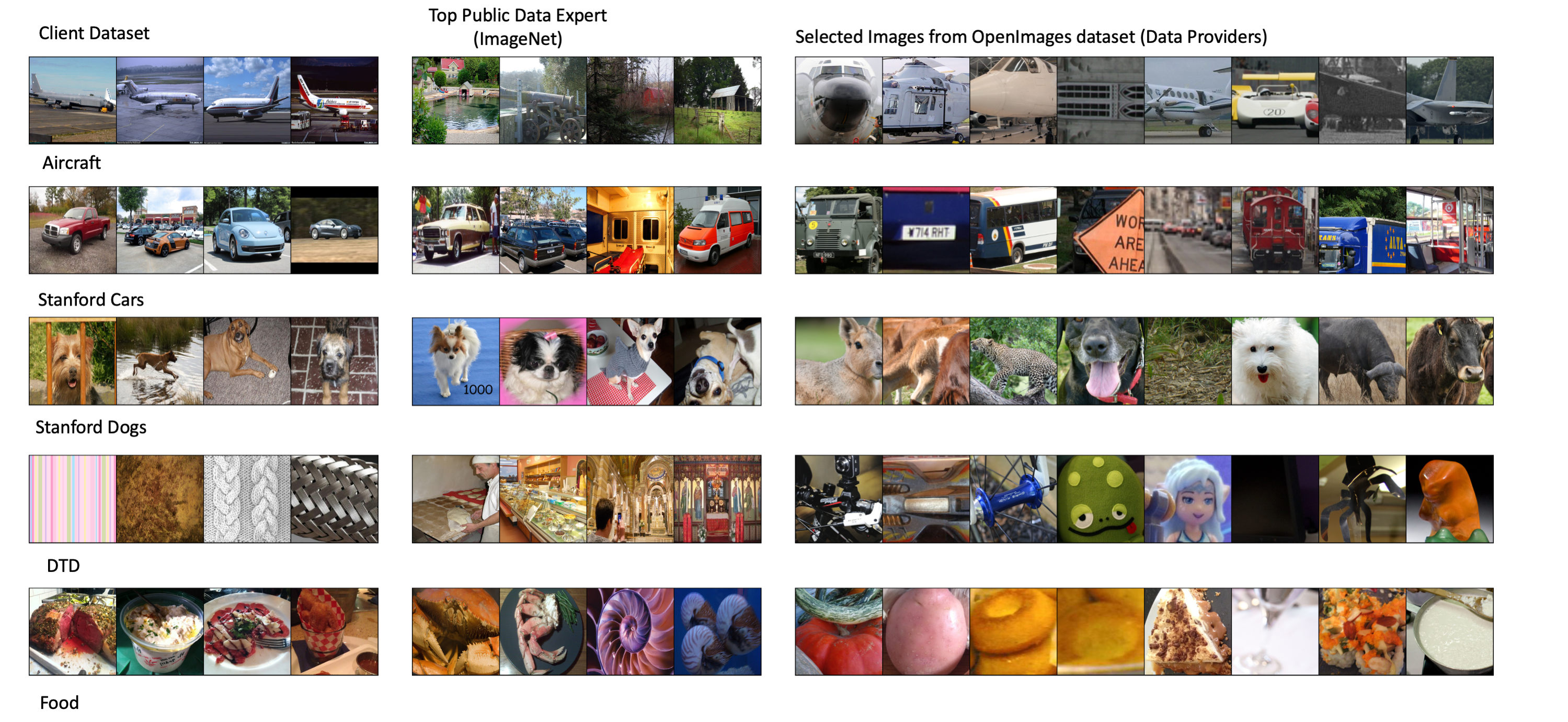}
    \vspace{-12mm}
    \includegraphics[width=0.97\textwidth]{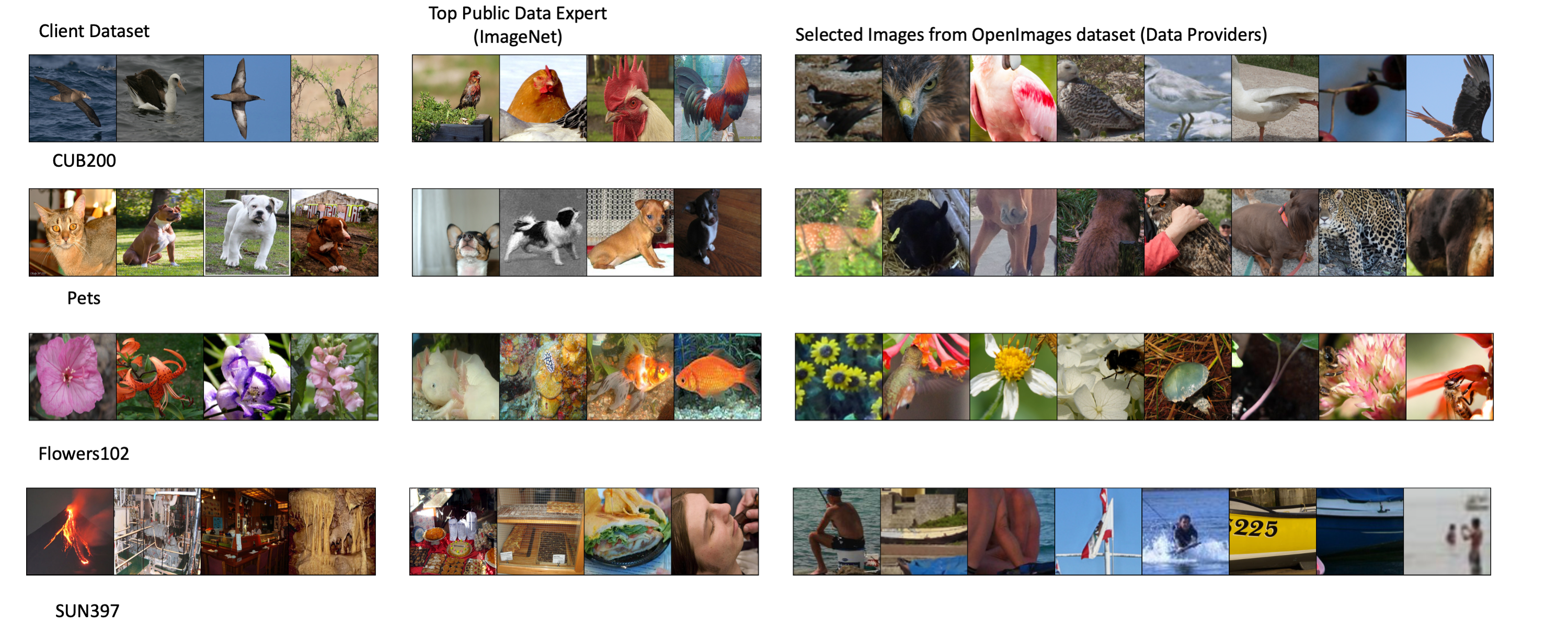}
    \vspace{12mm}
    \caption{\small %Examples of selected source images from OpenImages and images from ``most-similar public split" from ImageNet for each client dataset, and the associated top classes for the sampled distribution.
    From left to right: Images from client/consumer dataset, images from ``most similar" public split (ImageNet), selected images from data sources (Openimages).
   }
    \label{fig:supp_selected}
%    \vspace{-15pt}
%\end{wrapfigure}
\vspace{-10mm}
\end{figure}

\clearpage
\begin{figure}[t!]
    \vspace{-10mm}
    \centering
    \includegraphics[width=0.99\textwidth]{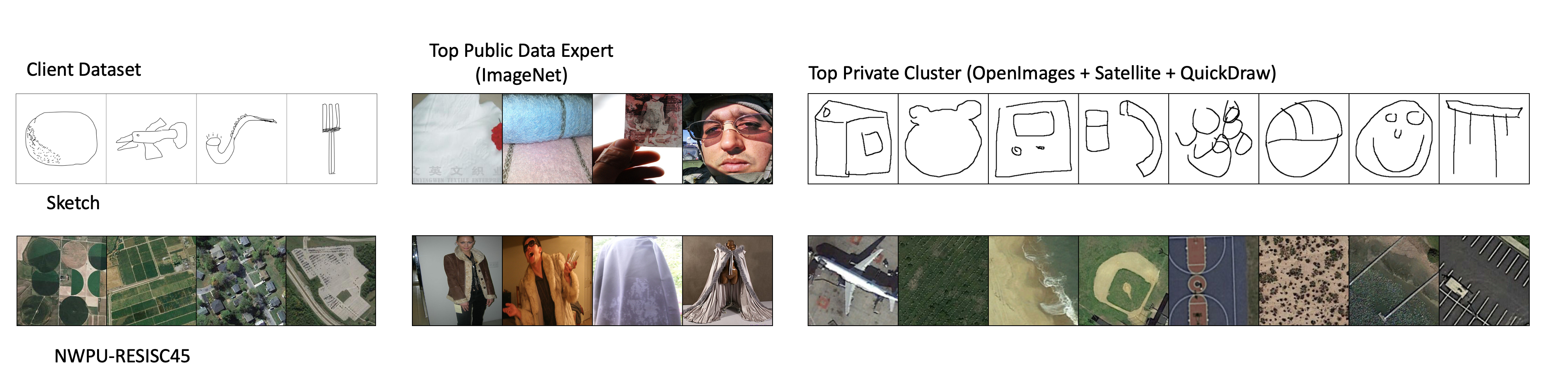}
    \vspace{-3mm}
    \includegraphics[width=0.99\textwidth]{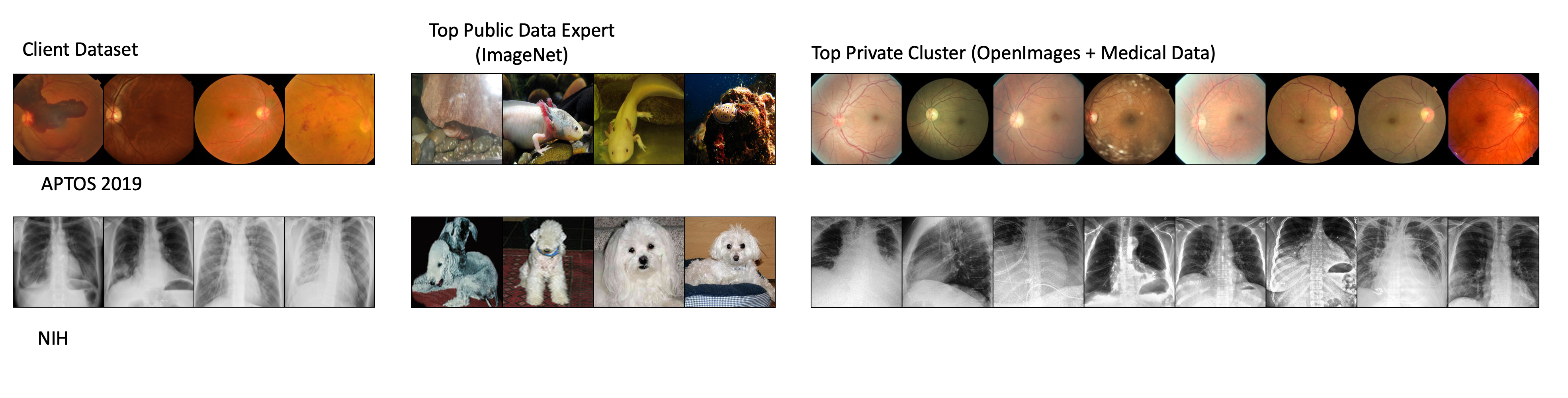}
    \caption{\small %Examples of selected source images from OpenImages and images from ``most-similar public split" from ImageNet for each client dataset, and the associated top classes for the sampled distribution.
    From left to right: Images from client/consumer dataset, images from ``most similar" public split (ImageNet), Top cluster in corresponding mixed source dataset.
   }
    \label{fig:supp_dissimilar}
%    \vspace{-15pt}
%\end{wrapfigure}
% \vspace{-8mm}
\end{figure}

\end{document}